\documentclass{article}
\usepackage[margin=1in]{geometry}
\usepackage{times}
\usepackage[authoryear,round]{natbib}
\setcitestyle{citesep={;},aysep={,},yysep={;}}
\usepackage[T1]{fontenc}
\usepackage[utf8]{inputenc}
\usepackage{amsmath,amssymb,amsthm,mathtools,booktabs,graphicx,microtype,xcolor}
\usepackage{algorithm,algorithmic}
\usepackage{hyperref,url}
\hypersetup{hidelinks}

\newcommand{\E}{\mathbb E}

\newsavebox{\FitBox}
\newcommand{\fitwidth}[1]{%
  \sbox{\FitBox}{#1}%
  \ifdim\wd\FitBox>\linewidth
    \typeout{TABLEFIT: natural=\the\wd\FitBox; limit=\the\linewidth; shrink}%
    \resizebox{0.9\linewidth}{!}{\usebox{\FitBox}}%
  \else
    \typeout{TABLEFIT: natural=\the\wd\FitBox; limit=\the\linewidth; native}%
    \usebox{\FitBox}%
  \fi}
  
\newcommand{\fitwidthfull}[1]{%
  \sbox{\FitBox}{#1}%
  \ifdim\wd\FitBox>\linewidth
    \typeout{TABLEFIT: natural=\the\wd\FitBox; limit=\the\linewidth; shrink}%
    \resizebox{\linewidth}{!}{\usebox{\FitBox}}%
  \else
    \typeout{TABLEFIT: natural=\the\wd\FitBox; limit=\the\linewidth; native}%
    \usebox{\FitBox}%
  \fi}
  
\newtheorem{theorem}{Theorem}
\newtheorem{lemma}[theorem]{Lemma}
\newtheorem{corollary}[theorem]{Corollary}

\title{Implicit Credit Learning for Sparse-Reward\\Multi-Agent Reinforcement Learning}
\author{The Viet Bui \and Tien Mai \and Hong Thanh Nguyen}
\usepackage{xcolor}

\begin{document}
\date{Working paper}
\maketitle

\newcommand{\IMAPSmacTable}{%
\begin{table}[t]
\centering
\caption{SMACv2 win rate (\%). Mean $\pm$ standard deviation over five seeds after 10M environment steps. All methods receive only terminal win/loss feedback. IMAP-LA and IMAP-GA denote the reported local- and global-advantage configurations with the same Q/V reward-model design.}
\label{tab:smacv2_results}
\fitwidthfull{%
\begin{tabular}{lcccccc}
\toprule
Scenario & SparseMAPPO & SparseHAPPO & Online-IPL & SL-MAPPO & IMAP-GA & IMAP-LA \\
\midrule
protoss\_5\_vs\_5 & $0.00 \pm 0.00$ & $11.61 \pm 3.03$ & $15.03 \pm 3.80$ & $26.97 \pm 6.00$ & $43.52 \pm 2.21$ & $\mathbf{44.00 \pm 3.16}$ \\
protoss\_10\_vs\_10 & $0.00 \pm 0.00$ & $0.00 \pm 0.00$ & $9.35 \pm 2.50$ & $21.38 \pm 3.58$ & $31.71 \pm 2.19$ & $\mathbf{33.64 \pm 2.78}$ \\
protoss\_10\_vs\_11 & $0.00 \pm 0.00$ & $0.00 \pm 0.00$ & $7.54 \pm 2.33$ & $15.67 \pm 3.48$ & $22.01 \pm 3.75$ & $\mathbf{23.08 \pm 3.99}$ \\
terran\_5\_vs\_5 & $31.96 \pm 3.26$ & $35.46 \pm 3.58$ & $32.47 \pm 4.42$ & $38.11 \pm 5.57$ & $45.26 \pm 2.15$ & $\mathbf{46.84 \pm 2.93}$ \\
terran\_10\_vs\_10 & $23.12 \pm 4.24$ & $23.47 \pm 2.85$ & $23.40 \pm 2.66$ & $28.22 \pm 4.16$ & $36.23 \pm 3.26$ & $\mathbf{38.18 \pm 2.82}$ \\
terran\_10\_vs\_11 & $0.00 \pm 0.00$ & $0.00 \pm 0.00$ & $10.65 \pm 1.78$ & $14.72 \pm 2.45$ & $17.46 \pm 4.25$ & $\mathbf{19.71 \pm 2.08}$ \\
zerg\_5\_vs\_5 & $18.45 \pm 3.99$ & $23.01 \pm 3.92$ & $28.23 \pm 3.46$ & $25.80 \pm 3.74$ & $38.48 \pm 5.48$ & $\mathbf{42.14 \pm 3.44}$ \\
zerg\_10\_vs\_10 & $10.40 \pm 2.54$ & $13.26 \pm 2.02$ & $14.40 \pm 2.26$ & $20.98 \pm 1.37$ & $26.15 \pm 1.51$ & $\mathbf{27.78 \pm 3.65}$ \\
zerg\_10\_vs\_11 & $4.63 \pm 1.63$ & $4.76 \pm 1.60$ & $4.78 \pm 1.91$ & $8.73 \pm 2.72$ & $14.42 \pm 2.80$ & $\mathbf{17.24 \pm 1.77}$ \\
\bottomrule
\end{tabular}
}
\end{table}
}

\newcommand{\IMAPMujocoTable}{%
\begin{table}[t]
\centering
\caption{MAMuJoCo episode return. Mean $\pm$ standard deviation over five seeds after 10M environment steps. Each learner receives the cumulative environment return only at episode termination.}
\label{tab:mamujoco_results}
\fitwidth{%
\begin{tabular}{lcccc}
\toprule
Method & Hopper-v2 & Reacher-v2 & HalfCheetah-v2 & Swimmer-v2 \\
\midrule
SparseMAPPO & $188.57 \pm 18.34$ & $-488.88 \pm 76.07$ & $1824.33 \pm 181.46$ & $16.29 \pm 2.12$ \\
SparseHAPPO & $215.41 \pm 30.01$ & $-497.58 \pm 77.65$ & $1926.69 \pm 224.82$ & $20.54 \pm 2.89$ \\
Online-IPL & $243.51 \pm 9.67$ & $-359.19 \pm 17.31$ & $1777.58 \pm 341.44$ & $23.10 \pm 0.81$ \\
SL-MAPPO & $296.45 \pm 19.09$ & $-358.15 \pm 15.53$ & $2479.37 \pm 337.53$ & $29.84 \pm 1.21$ \\
IMAP-GA & $296.48 \pm 27.55$ & $-316.83 \pm 67.58$ & $3328.71 \pm 285.84$ & $31.89 \pm 1.18$ \\
IMAP-LA & $\mathbf{361.22 \pm 16.19}$ & $\mathbf{-182.10 \pm 13.25}$ & $\mathbf{3853.58 \pm 359.02}$ & $\mathbf{33.98 \pm 1.04}$ \\
\bottomrule
\end{tabular}
}
\end{table}
}

\newcommand{\IMAPSmacLiteTable}{%
\begin{table}[t]
\centering
\caption{SMACLite terminal-feedback returns (five-seed mean $\pm$ SD).}
\label{tab:smaclite_sota}
\fitwidth{%
\begin{tabular}{lccccc}
\toprule
Scenario & SparseMAPPO & AREL & STAS & TAR$^2$ & IMAP-LA \\
\midrule
2s\_vs\_1sc & $3.43 \pm 0.93$ & $9.40 \pm 0.13$ & $9.95 \pm 0.02$ & $13.24 \pm 0.96$ & $\mathbf{13.64 \pm 0.85}$ \\
3s\_vs\_5z & $4.42 \pm 0.05$ & $7.61 \pm 0.50$ & $9.22 \pm 0.89$ & $9.60 \pm 0.85$ & $\mathbf{9.98 \pm 0.81}$ \\
3s5z & $8.45 \pm 2.60$ & $19.15 \pm 0.68$ & $19.82 \pm 0.19$ & $\mathbf{19.91 \pm 0.16}$ & $19.75 \pm 0.26$ \\
\bottomrule
\end{tabular}
}
\end{table}
}

\newcommand{\IMAPHeldoutTable}{%
\begin{table}[t]
\centering
\caption{Held-out reward-model diagnostics. Both frozen models score the same 2000 pairs from a shared pool of 256 trajectories per task and seed after 10M training steps. Values are descriptive means over three seeds.}
\label{tab:heldout_reward}
\fitwidth{%
\small
\begin{tabular}{lcccc}
\toprule
& \multicolumn{2}{c}{Ranking accuracy $\uparrow$} & \multicolumn{2}{c}{BT NLL $\downarrow$}\\
\cmidrule(lr){2-3}\cmidrule(lr){4-5}
Task & IMAP-GA & SL-MAPPO & IMAP-GA & SL-MAPPO\\
\midrule
protoss\_5\_vs\_5 & 0.857 & 0.841 & 0.618 & 0.632\\
terran\_5\_vs\_5 & 0.851 & 0.828 & 0.605 & 0.654\\
Hopper-v2 & 0.891 & 0.812 & 0.517 & 0.634\\
\bottomrule
\end{tabular}}
\end{table}}

\begin{abstract}
When a cooperative team receives feedback only at the end of an episode, a single outcome must guide many agents' earlier actions. Learning a dense team reward helps, but its decomposition does not automatically give each agent a useful policy update. We introduce IMAP, an online method that fits a shared Q/V reward model by comparing completed episodes according to their observed terminal outcomes. Local Q/V residuals provide agent-specific rewards; the same normalized weights combine them for a centralized critic and scale local advantages for decentralized PPO actors. We characterize when these local updates follow the \emph{learned} team objective. With full returns and frozen reward heads, a sharp bound relates their gradient error to variation in future Q/V compatibility; an exact identity identifies additional continuation terms for general GAE. In experiments with terminal-only feedback, IMAP's local-actor configuration attains the highest reported mean among the compared methods on all nine SMACv2 scenarios and four MAMuJoCo tasks, and on two of three SMACLite scenarios. Controlled relay interventions confirm the predicted changes in actor credit while holding the learned team reward fixed. The gradient result concerns the fitted objective; benchmark returns measure performance on the original task outcomes.
\end{abstract}

\section{Introduction}\label{sec:intro}
In many cooperative tasks, agents act for a long time before the team receives a single outcome: a battle is won or lost, or a coordinated motion earns an episode return. That outcome identifies success, but gives no direct indication of which agent's action mattered or when. The problem is therefore both temporal and multi-agent credit assignment. It matters especially under centralized training with decentralized execution: training may use team information, but each deployed actor must choose actions from its own observations.

A learned dense reward can distribute terminal feedback across time, yet a team-level advantage still supplies every actor with the same reward signal. Decomposing that reward can provide local signals, but decomposition alone is insufficient. An agent's action may change the future states and rewards of other agents, so optimizing only its own component can depart from the learned team objective. Prior shared Q/V reward fitting uses residual comparisons and offline weighted policy extraction \citep{bui2025omapl}. The missing connection here is an online, terminal-outcome-driven local PPO update, together with a characterization of when its local credit tracks the team objective.

IMAP makes that connection with one reward model and one mixture. Completed episodes are ranked by their observed terminal outcomes to fit shared Q/V heads; no external preference labels are needed. Each head yields an inverse-Bellman residual that acts as a local reward. Normalized weights combine these residuals into the reward for a centralized critic and scale the corresponding advantages for decentralized PPO actors. Thus the critic and actors learn from the same fitted team score, while the actors receive distinct local credit. The method works with the observed win/loss signal or a cumulative return revealed only at episode end.

The central analytical question is whether distinct actor signals still point in the direction of the \emph{learned} team return. The mixture makes global and weighted local advantages equal as numerical sums, but equality of those sums does not guarantee equality of their policy gradients: an action can affect other agents' future credit. We therefore study the local direction under frozen reward heads and evaluate task performance separately against the observed terminal outcome.

Our contributions are:
\begin{itemize}
\item \textbf{An online local-actor algorithm for terminal feedback.} IMAP turns comparisons among completed episodes into Q/V residual rewards, then uses one normalized mixture for a global critic and weighted local PPO updates. This connects shared Q/V fitting to online decentralized policy improvement, whereas the prior Q/V approach uses offline weighted policy extraction \citep{bui2025omapl}.
\item \textbf{A precise account of when local credit follows the team score.} An exact identity isolates the continuation terms at general GAE. For full returns, a sharp trajectory-Fisher bound reduces the error to future Q/V compatibility without an explicit agent-count factor; a separate read-once result lower-bounds ideal local signal in structured coordination. The gradient guarantee concerns the fitted objective, with conditions for transfer to the true outcome given in Appendix~\ref{app:terminal-transfer}.
\item \textbf{Evidence for both performance and the credit mechanism.} Under terminal-only feedback, IMAP's local-actor configuration has the highest reported mean among the compared methods on all nine SMACv2 scenarios and four MAMuJoCo tasks, and on two of three SMACLite scenarios. Model and actor-readout comparisons probe the design choices, while frozen relay interventions test the predicted credit changes at unchanged mixed reward.
\end{itemize}

\section{Related work}\label{sec:related}
\paragraph{Multi-agent policy optimization.}
CTDE enables agents to learn from global information during training while acting independently at execution \citep{foerster2018counterfactual,lowe2017multi,kraemer2016multi}. Value decomposition supplies factorized representations of a team objective \citep{sunehag2017value,rashid2020monotonic,son2019qtran,wang2020qplex}.
MAPPO and HAPPO develop cooperative PPO updates \citep{schulman2017proximal,yu2022surprising_MAPPO,kuba2021trust}. DOP derives weighted local policy gradients from a decomposed critic \citep{wang2021dop}; IMAP analyzes the temporal credit induced by learned Q/V residual rewards. COMA constructs a counterfactual advantage using a centralized Q-function \citep{foerster2018counterfactual}. IMAP instead obtains its local signals from a reward model fitted to terminal outcomes. The present analysis states when these signals reproduce the learned-team gradient and bounds their discrepancy otherwise.

\paragraph{Learning from delayed rewards.}
Reward redistribution converts episodic feedback into denser signals \citep{arjona2019rudder,ren2021learning}.
AREL, STAS and TAR$^2$ learn temporal or agent contributions \citep{xiao2022agent,chen2024stas,kapoor2025redistributingrewardstimeagents}.
DIFFER decomposes rewards for individual replay, LAIES uses model-based causal intrinsic rewards, and MASER generates replay-derived subgoals \citep{hu2023differ,liu2023lazy,jeon2022maser}.
The latter three emphasize replay or exploration; IMAP studies local actor credit from terminal-outcome-fitted Q/V rewards.

\paragraph{Inverse value learning and actor geometry.}
The inverse-Bellman likelihood and exponential value update follow IPL/XQL \citep{hejna2024inverse,garg2023extreme}, building on inverse Q-learning \citep{garg2021iq}. O-MAPL already combines shared multi-agent Q/V fitting, residual comparisons and Extreme-V with offline weighted policy extraction \citep{bui2025omapl}. IMAP specializes to normalized scalar weights, online terminal feedback and local PPO. Advantage-support analysis expresses bias through omitted score--reward couplings \citep{zhao2026aggregate}; our temporal analysis reduces them to future component compatibility and gives sharp trajectory-Fisher control, including shared parameters \citep{kakade2001natural}.

\section{Cooperative learning from terminal outcomes}\label{sec:setup}
We focus on cooperative MARL, modeled as a multi-agent partially observable Markov decision process. The team has agents $\mathcal N=\{1,\ldots,n\}$ and joint action space $\mathcal A=\prod_i\mathcal A_i$.
At stage $t$, the environment draws the next state from $P(\cdot\mid \mathbf{s}_t,\mathbf a_t)$. Agent $i$ acts from its observation or history $o_{i,t}$, and the joint policy factorizes as $\pi_\theta(\mathbf a_t\mid \mathbf{s}_t)=\prod_i\pi_{i,\theta}(a_{i,t}\mid o_{i,t})$. Here $\mathbf{s}_t$ is a sufficient information state for analysis; it can contain the joint history. The actors remain decentralized. Their parameters may be shared.

A complete episode $\tau$ supplies one terminal team outcome $G^*(\tau)$. The learning objective is $J^*(\pi)=\mathbb E_\pi G^*(\tau)$. In the binary battle tasks, this outcome is the terminal win/loss signal. In the continuous-control tasks, the environment's cumulative episode return is withheld until termination. The learner never receives those intermediate task rewards during training. Evaluation uses the original team outcome in its original units.

The fitted model assigns each episode a discounted score $\widehat F(\tau)$, defined below from its per-step rewards. Its expectation $J_{\widehat R}=\E_\pi\widehat F$ is the \emph{learned} team objective used in our credit analysis. This differs from the observed-outcome objective $J^*$ unless additional calibration conditions hold; Appendix~\ref{app:terminal-transfer} states those conditions.

\section{Implicit rewards and local policy updates}\label{sec:method}
\begin{figure}[t]
\centering
\includegraphics[width=\linewidth]{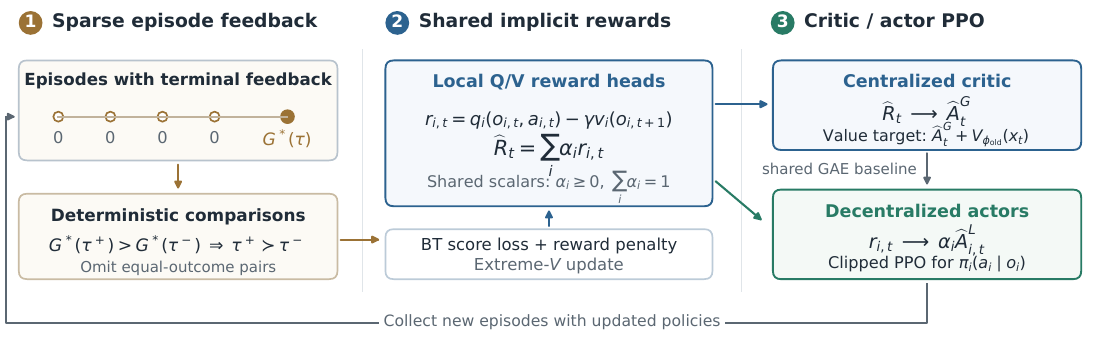}
\caption{IMAP learns from sparse team outcomes. Completed episodes train a shared Q/V reward model; its local residuals supply a mixed reward for the centralized critic and agent-specific advantages for decentralized actors. Outcome comparisons are computed from the observed terminal returns.}
\label{fig:imap_overview}
\end{figure}

Comparing terminal outcomes can fit a team-level trajectory score, but it does not tell a PPO actor which of its own actions to reinforce. Reading local terms from that score is also delicate: an action can change other agents' future rewards. IMAP's methodological contribution is a closed online loop. The current policy generates complete episodes; their observed outcomes label comparisons; fitted Q/V residuals supply dense rewards for the next PPO update. One normalized mixture defines the critic's team reward and weights the actors' local advantages. Unlike O-MAPL's offline preference learning and weighted behavior-cloning extraction \citep{bui2025omapl}, this loop updates the data-generating policy with local credit. Section~\ref{sec:theory} characterizes when that credit follows the learned team objective; Figure~\ref{fig:imap_overview} depicts the loop.

\subsection{A shared Q/V mixture for team and local rewards}
Additive value decomposition is established in cooperative MARL: VDN sums local values, DOP uses a weighted linear critic for decomposed policy gradients, and O-MAPL applies linear Q/V mixing to preference-based learning \citep{sunehag2017value,wang2021dop,bui2025omapl}. We use a normalized version because the critic and local PPO actors must read from the same fitted score. The local heads $q_i(o_i,a_i)$ and $v_i(o_i)$ have separate reward-model parameters. Their common mixer is
\begin{align}
Q_{\rm tot}(\mathbf{s},\mathbf a)&=\sum_i\alpha_i q_i(o_i,a_i), &
V_{\rm tot}(\mathbf{s})&=\sum_i\alpha_i v_i(o_i),\label{eq:mixer}\\
\alpha_i&=\frac{\operatorname{softplus}(\omega_i)}{\sum_j\operatorname{softplus}(\omega_j)}.
\end{align}
The nonnegative weights sum to one and are shared by Q and V. They are global scalars, with no additive bias. Thus the mixed reward is exactly the weighted sum of local Q/V residuals, while normalization preserves the common critic term in GAE (Equation~\eqref{eq:gae-sum}). Using these same $\alpha_i$ in local PPO ties each actor to its contribution to the fitted team score. This tractable class excludes state-dependent mixing; our experiments do not compare mixer classes. The heads and weights are frozen during each actor update.

For a sampled transition $(\mathbf{s}_t,\mathbf a_t,\mathbf{s}_{t+1})$, define
\begin{equation}
r_{i,t}=q_i(o_{i,t},a_{i,t})-\gamma v_i(o_{i,t+1}),\qquad
\widehat R_t=\sum_i\alpha_i r_{i,t}.
\label{eq:implicit-reward}
\end{equation}
Here $\gamma$ is the discount factor. The next-observation value uses the successor reached in the rollout, and terminal values are zero. We call $r_{i,t}$ an \emph{implicit} reward because it is computed from Q/V heads rather than predicted by a separate reward head. Averaging over transitions gives the expected-transition inverse-Bellman residual.

\subsection{Learning dense rewards from terminal outcomes}
The environment supplies no per-step training reward, but completed episodes can be ordered by their observed outcomes. For two episodes $\tau_1,\tau_2$, define $\tau_1\succ\tau_2$ if $G^*(\tau_1)>G^*(\tau_2)$; discard ties. These labels come from online interaction, without an external preference source.
Let $\widehat F(\tau)=\sum_t\gamma^t\widehat R_t$. The BT pair model and reward-fitting objective are
\begin{align}
p(\tau^+\succ\tau^-)&=\operatorname{sigmoid}\big(\widehat F(\tau^+)-\widehat F(\tau^-)\big),\label{eq:bt}\\
\mathcal L_Q&=-\mathbb E_{\mathcal P}\log p(\tau^+\succ\tau^-)
+\lambda_R\mathbb E_{\mathcal D}\widehat R_t^2.\label{eq:rewardloss}
\end{align}
Here $\mathcal P$ denotes outcome-ordered episode pairs and $\mathcal D$ stored transitions; $\lambda_R$ weights a squared-reward regularizer. The BT loss trains a trajectory score from sparse outcomes, while the regularizer controls its scale. Ordering discards continuous-return magnitudes, so good ranking does not establish calibration to $G^*$. Q-head and mixer updates use stored episodes with V-heads fixed.

The V-heads complete the residual reward: fitting episode rankings through Q alone would leave its next-state subtraction unspecified. Following Extreme-$V$ \citep{garg2023extreme}, we encourage the mixed V value to track a soft value of the mixed Q-function:
\begin{equation}
\mathcal L_V=\mathbb E_{(\mathbf{s},\mathbf a)\sim\mathcal D}
\left[\exp\!\left(\frac{Q_{\rm tot}(\mathbf{s},\mathbf a)-V_{\rm tot}(\mathbf{s})}{\beta}\right)
-\frac{Q_{\rm tot}(\mathbf{s},\mathbf a)-V_{\rm tot}(\mathbf{s})}{\beta}-1\right].
\label{eq:extremev}
\end{equation}
The temperature is $\beta>0$. The V loss uses buffered actions; Section~\ref{sec:theory} instead measures each V-head's mismatch with its Q-head's current-policy mean at frozen checkpoints. Fitting the mixed V need not eliminate that mismatch. The PPO critic is trained separately.

\subsection{Global evaluation with weighted local PPO}
MAPPO gives every actor the same mixed-reward advantage \citep{yu2022surprising_MAPPO}. IMAP gives actors distinct residual-based advantages while a centralized critic $V_\phi(\mathbf{s}_t)$ evaluates the team reward. This may sharpen sparse-feedback credit; Section~\ref{sec:experiments} compares the two readouts. Each actor $\pi_{i,\theta}$ uses only its local observation or history.
Fix its advantage-computation snapshot $V_{\phi_{\rm old}}$ and write
\begin{align}
B_t&=\gamma V_{\phi_{\rm old}}(\mathbf{s}_{t+1})-V_{\phi_{\rm old}}(\mathbf{s}_t),\nonumber\\
\delta^G_t&=\widehat R_t+B_t,\qquad \delta^L_{i,t}=r_{i,t}+B_t,\nonumber\\
\widehat A^G_t&=\sum_{\ell\ge0}(\gamma\lambda)^\ell\delta^G_{t+\ell},\qquad
\widehat A^L_{i,t}=\sum_{\ell\ge0}(\gamma\lambda)^\ell\delta^L_{i,t+\ell}.\label{eq:gae}
\end{align}
Here $\lambda$ is the GAE trace parameter. Both streams use the same terminal or rollout masks and bootstrap convention; on complete episodes they stop at the true terminal boundary. Because the mixer weights sum to one, the sampled advantages obey the exact identity
\begin{equation}
\widehat A^G_t=\sum_i\alpha_i\widehat A^L_{i,t}.
\label{eq:gae-sum}
\end{equation}
Mixing the local TD residuals gives the global residual, and the common GAE sum preserves that equality. The identity is an implementation check, but it does not imply that local and global actor gradients agree: each actor multiplies its own score by a different advantage. Section~\ref{sec:theory} analyzes the difference.

For each agent define the individual likelihood ratio
$\rho_{i,t}(\theta)=\pi_{i,\theta}(a_{i,t}\mid o_{i,t})/\pi_{i,\theta_{\rm old}}(a_{i,t}\mid o_{i,t})$.
The IMAP actor objective is
\begin{equation}
\mathcal L_{\rm actor}(\theta)=\mathbb E_t\sum_i\left[
\alpha_i\min\!\left\{\rho_{i,t}\widehat A^L_{i,t},
\operatorname{clip}(\rho_{i,t},1-\epsilon_{\rm PPO},1+\epsilon_{\rm PPO})\widehat A^L_{i,t}\right\}
+\eta\mathcal H(\pi_{i,\theta})\right].
\label{eq:actor}
\end{equation}
Here $\epsilon_{\rm PPO}$ is the clipping width, $\eta$ is the entropy coefficient, and $\mathcal H$ denotes policy entropy. The ratio uses the policy snapshot $\theta_{\rm old}$ that collected the rollout.
The coefficient $\alpha_i$ is the component's weight in the fitted team reward, so it also scales that actor's PPO reward term, as in decomposed policy gradients \citep{wang2021dop}. Global controls use $\widehat A_t^G$ with a unit prefactor; both variants retain the same entropy term. Here $\E_t$ averages rollout actor samples. For a one-stage episode with frozen heads and critic, conditional action independence gives the expected old-policy reward-gradients
\begin{equation}
g_G=\E\sum_i z_i\Big(\sum_j\alpha_jq_j-V\Big)
=\sum_i\alpha_i\E[z_iq_i]=g_L,
\label{eq:actor-coefficients}
\end{equation}
where $z_i=\nabla_\theta\log\pi_i$, $V$ is the shared frozen baseline, and $g_L=\E\sum_i\alpha_i z_i(q_i-V)$.
This equality concerns the expected old-policy reward gradient for one stage, not the full clipped PPO update. Appendix~\ref{app:objective-consistency} gives the shared-parameter proof.
Section~\ref{sec:theory} characterizes the sequential discrepancy.

The critic uses the global target $\widehat Y_t=\widehat A^G_t+V_{\phi_{\rm old}}(\mathbf{s}_t)$ and MAPPO's clipped value loss. Reward-model parameters and advantage targets remain fixed during PPO epochs; critic updates do not change these targets.

\begin{algorithm}[t]
\caption{IMAP with terminal-outcome feedback}
\label{alg:imap}
\begin{algorithmic}[1]
\STATE Initialize actors, critic, Q/V heads, normalized mixer, and both buffers.
\FOR{each iteration}
\STATE Collect rollouts and append completed episodes with their terminal outcomes.
\STATE Fit Q/mixer by \eqref{eq:rewardloss} and V by \eqref{eq:extremev} using available unequal-return pairs.
\STATE Freeze model/critic snapshots; compute both GAE streams with common masks.
\STATE Run PPO epochs using local actor targets and global critic targets.
\STATE Clear the on-policy rollout buffer.
\ENDFOR
\end{algorithmic}
\end{algorithm}

Common affine reward normalization and masks preserve the mixture identities. Appendix~\ref{app:implementation} specifies empirical reductions and normalization; Appendix~\ref{app:bootstrap} gives bootstrap corrections.

\section{When local credit follows the team objective}\label{sec:theory}
Equation~\eqref{eq:actor-coefficients} explains the local coefficient for one decision. Over an episode, an agent's action can alter the states in which other agents later act. Their future reward components can then change even when the current local term is correctly weighted. We first identify this continuation error at the GAE trace parameter and then bound it sharply for complete returns.

\paragraph{Setting.}
Consider an $H$-stage Markov game with discount $\gamma\in[0,1]$ and a
sufficient pre-action state $\mathbf{S}_h$, which may contain joint histories.
Given $\mathbf{S}_h$, the agents sample actions independently from their local
policies. Actor parameters may be shared. The initial law, transitions
and observations are independent of these parameters. Assume finite
state and action sets and differentiable policies positive on fixed
feasible supports; Appendix~\ref{app:regularity} covers general spaces.
Freeze the local heads, the critic $V_h$, and scalar weights
$\alpha_i\ge0$, $\sum_i\alpha_i=1$, with $v_{i,H}=V_H=0$.
Each Q-head uses only its own actor's action.

Define the learned-team objective and local actor vector by
\begin{align}
J_{\widehat R}(\theta)
&=\mathbb E_{\pi_\theta}\sum_{h=0}^{H-1}\gamma^h\widehat R_h,
\label{sf:actor-target}\\
g_L^{(\lambda)}
&=\mathbb E\sum_{h=0}^{H-1}\sum_i\gamma^h\alpha_i
z_{i,h}\widehat A^L_{i,h},\qquad
z_{i,h}=\nabla_\theta\log\pi_{i,\theta}(A_{i,h}\mid o_{i,h}).
\label{sf:actor}
\end{align}
Here $\nabla J_{\widehat R}:=\nabla_\theta J_{\widehat R}(\theta)$ differentiates the policy with the reward model fixed. The local vector is the reward-part derivative at the old policy with
objective-aligned time weights. Entropy, the empirical rollout reduction
and later clipped PPO steps are separate.

The key quantity is how well each V-head matches the current policy's mean Q-head. Define the \emph{compatibility error} as their difference:
\begin{equation}
e_{i,h}(\mathbf{s})=\mathbb E_{a_i\sim\pi_i(\cdot\mid o_i)}
q_{i,h}(o_i,a_i)-v_{i,h}(o_i),\qquad
\varepsilon=\max_{i,h<H}\operatorname{span}e_{i,h},
\label{sf:compatibility}
\end{equation}
where $\operatorname{span}f=\max_{\mathbf{s}}f(\mathbf{s})-\min_{\mathbf{s}}f(\mathbf{s})$. The bound below depends on variation across states, rather than on a constant offset between Q and V.
\subsection{Exact discrepancy under general GAE}
At finite trace length, the actor also depends on the gap between the centralized critic and its local V-head. Write $D_i=V-v_i$ and $\bar e=\sum_i\alpha_i e_i$.
For a sequence of state functions $f$, define its current-policy
continuation by
$(\mathcal R_c f)_h=\E[\sum_{t=h}^{H-1}c^{t-h}f_t(\mathbf{S}_t)\mid \mathbf{S}_h]$,
with value zero at $H$. Let $P_{i,h}^{a_i}$ average the next-state
kernel over the other agents' actions, and let $P_{\pi,h}$ average
over all actions. Define
\begin{equation}
F_i=\alpha_i\mathcal R_{\gamma\lambda}
\big((1-\lambda)D_i+\lambda e_i\big)-\mathcal R_\gamma\bar e.
\label{eq:continuation-main}
\end{equation}
For every $\lambda\in[0,1]$, the exact actor discrepancy is
\begin{equation}
g_L^{(\lambda)}-\nabla J_{\widehat R}
=\E\left[\sum_{h,i}\gamma^{h+1}z_{i,h}
\big[(P_{i,h}^{A_{i,h}}-P_{\pi,h})F_{i,h+1}\big](\mathbf{S}_h)\right].
\label{eq:gae-actor-error}
\end{equation}
Equation~\eqref{eq:gae-actor-error} measures how the acting agent changes the expected future quantity $F_i$ relative to the current policy. At training $\lambda=0.95$, the local continuation contains
$0.05D_i+0.95e_i$, together with the mixed-compatibility correction in
Equation~\eqref{eq:continuation-main}.
The identity permits shared actor parameters and arbitrary frozen heads.
Appendix~\ref{app:general-gae} proves it; Appendix~\ref{app:bootstrap}
gives the finite-rollout endpoint correction.

\subsection{Sharp control by future compatibility}

For full returns, the critic--head continuation term disappears. To express the remaining error in a scale that accounts for the policy's parameterization, define the trajectory score and its Fisher matrix:
\begin{equation}
S=\nabla_\theta\log p_\theta(\tau)=\sum_{h,i}z_{i,h},\qquad
\mathcal I_\pi=\mathbb E[SS^\top].
\label{sf:fisher}
\end{equation}
For $b\in\operatorname{range}(\mathcal I_\pi)$, let $\mathcal I_\pi^\dagger$ denote the Moore--Penrose pseudoinverse of the Fisher matrix in Equation~\eqref{sf:fisher}, allowing singular $\mathcal I_\pi$. The norm $\|b\|_{\mathcal I_\pi^\dagger}
=(b^\top\mathcal I_\pi^\dagger b)^{1/2}$ measures the largest error in
first-order predicted gain over directions of unit trajectory-Fisher
length.

\begin{theorem}[Sharp local-credit bound]\label{sf:theorem}
Under the preceding assumptions, full-return GAE ($\lambda=1$) satisfies
\begin{equation}
\boxed{\big\|g_L^{(1)}-\nabla J_{\widehat R}\big\|_{\mathcal I_\pi^\dagger}
\le\frac{\varepsilon}{2}\sum_{t=1}^{H-1}\gamma^t.}
\label{sf:bound}
\end{equation}
The discrepancy belongs to $\operatorname{range}(\mathcal I_\pi)$.
The constant $1/2$ is sharp uniformly over the allowed games and normalized
mixers: for $H\ge2$ and $\gamma>0$, two agents attain equality with a
nonnegative mixer, and strictly positive mixers approach equality.
\end{theorem}

At $\lambda=1$, Equation~\eqref{eq:continuation-main} reduces to
$F_i=-\mathcal R_\gamma\sum_{j\ne i}\alpha_j e_j$.
Thus the error comes from an actor's effect on other heads' future compatibility. The current-state terms cancel by conditional action independence; the shared critic acts as a baseline at $\lambda=1$. The theorem bounds the local direction relative to the learned reward objective, not directly relative to terminal-outcome performance. It applies to full returns, whereas the reported PPO runs use $\lambda=0.95$.
Consequently, state-constant compatibility errors give exact agreement,
even if the local Q/V values contain constant offsets.
More precisely, the proof bounds the discrepancy by
$\tfrac12\sum_{t=1}^{H-1}\gamma^t\max_i\operatorname{span}e_{i,t}$.
Only stages with varying compatibility contribute: a suffix with constant
compatibility adds no error term. Appendix~\ref{app:sharp-proof} proves the
bound and Appendix~\ref{app:sharpness} establishes sharpness.

\subsection{Available signal in structured coordination}\label{sec:signal-floor}
The preceding bound concerns credit accuracy, not the amount of local signal a task provides. We analyze the latter question in a restricted binary coordination class.

Consider independent binary decisions $A_i\sim\operatorname{Bernoulli}(p_i)$,
followed by an action-free continuation and terminal outcome $Y$.
Let $Y$ be a nonconstant monotone AND/OR expression in which each decision
occurs once: AND requires joint completion and OR allows alternative
successful routes. Such an expression is called \emph{read-once}.
A minimal successful coalition is a minimal set of actors whose activation
forces $Y=1$. Define the ideal local values and their signal energy by
\begin{equation}
q_i^\star(a)=\E[Y\mid A_i=a],\qquad
\mu=\E Y,\qquad
\mathcal C=\sum_i\operatorname{Var}(q_i^\star(A_i)).
\label{credit:energy-main}
\end{equation}
These are population quantities for the true outcome, distinct from fitted
reward-model heads.

\begin{theorem}[Sharp terminal-credit floor]\label{credit:signal-main}
In the read-once model above, suppose every minimal successful coalition has at most $k$ actors,
$0<\rho\le1/2$, $\rho\le p_i<1$, and $0<\mu\le1/2$. Then
\begin{equation}
\boxed{\mathcal C\ge a_k(\rho)(1-\mu)^2\log\frac1{1-\mu},\qquad
a_k(\rho)=\frac{k\rho^{2k-1}(1-\rho)}
{(1-\rho^k)^2\log\frac1{1-\rho^k}}.}
\label{credit:floor-main}
\end{equation}
Equality holds precisely for disjoint size-$k$ successful coalitions,
with every relevant probability equal to $\rho$.
\end{theorem}

Within this class, the floor depends on success probability, coalition width and exploration, with no additional team-size factor. It concerns the true outcome, not fitted reward heads. Appendix~\ref{credit:signal-appendix} proves the theorem and a single-cross-coalition extension; Appendix~\ref{app:signal-transfer} gives the calibrated transfer condition.

\section{Experiments}\label{sec:experiments}
The experiments ask three questions: how well the complete learner performs with terminal-only feedback, whether the Q/V model and local actor readout each contribute, and whether the credit errors predicted by the analysis appear when team reward is held fixed. Benchmarks address the first two; frozen-model relay interventions address the third.

\subsection{Benchmark setup and comparisons}
SMACv2 covers nine scenarios across Protoss, Terran and Zerg, with shaped rewards removed and terminal $+1/-1$ win/loss feedback \citep{ellis2023smacv2}. MAMuJoCo withholds per-step rewards and reveals the cumulative episode return at termination \citep{de2020deep_mamujoco}. Its agent partitions are $3\!\times\!1$ for Hopper, $2\!\times\!1$ for Reacher, $6\!\times\!1$ for HalfCheetah and $2\!\times\!1$ for Swimmer, with one-hop observations and an episode limit of 1000.

Each method uses 10M steps and five seeds. Evaluation samples 32 stochastic-policy episodes every 50k steps; tables average the final five checkpoints per seed and report the seed mean and standard deviation. PPO uses $\gamma=0.99$, $\lambda=0.95$, clipping $0.2$ and ten epochs. Architectures and budgets are in Appendix~\ref{app:implementation}.

SparseMAPPO and SparseHAPPO optimize terminal rewards directly. Online-IPL extracts policies from learned Q/V values by advantage-weighted behavior cloning. SL-MAPPO fits a direct reward head with the same terminal-outcome comparisons and uses global advantages. IMAP-GA uses the Q/V reward model with a global readout; IMAP-LA uses weighted local advantages and is the original IMAP-Rule configuration.

The SL-MAPPO/IMAP-GA comparison probes the Q/V construction against a direct reward head under the shared buffer, label-bearing update schedule and encoder design. The IMAP-GA/IMAP-LA comparison probes the actor readout. Each configuration collects its own on-policy data, so these are training-run comparisons rather than interventions on an identical fixed dataset; Appendix~\ref{app:implementation} specifies their scope.

\IMAPSmacTable
\subsection{Learning from terminal team feedback}
IMAP-LA achieves the highest mean win rate on all nine SMACv2 scenarios. The Q/V construction improves the global-readout configuration over the direct reward head on every scenario: on Protoss $10$ versus $10$, the mean increases from $21.38\%$ for SL-MAPPO to $31.71\%$ for IMAP-GA. The local-actor configuration further improves all nine means, by $0.48$--$3.66$ percentage points. Table~\ref{tab:smacv2_results} reports variability over seeds.

\begin{figure}[t]
\centering
\includegraphics[width=0.9\linewidth]{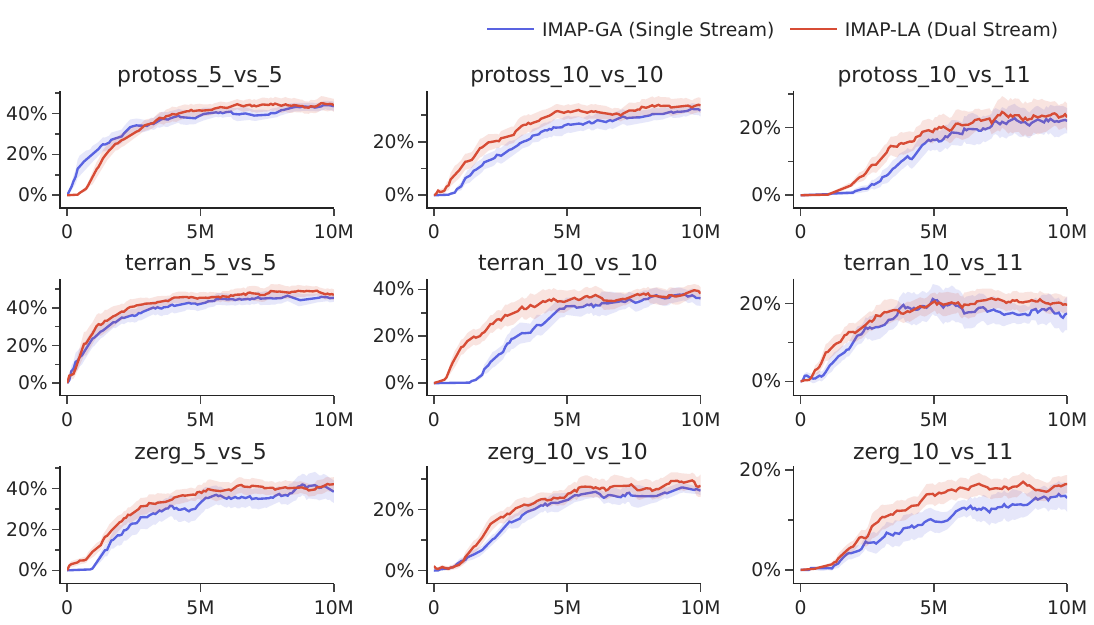}
\caption{Global and weighted local actor readouts on SMACv2. Both use the same Q/V reward-model design and a global critic; curves show win rate over 10M steps.}
\label{fig:smac_adv}
\end{figure}
\IMAPMujocoTable
The same ordering of reported means holds across the four continuous-control tasks. On Hopper, the two global-readout configurations are close ($296.45$ and $296.48$), while IMAP-LA reaches $361.22$. On HalfCheetah, the mean rises from $2479.37$ for SL-MAPPO to $3328.71$ for IMAP-GA and $3853.58$ for IMAP-LA. These comparisons support the combined design across discrete and continuous action spaces, while their seed variability and separately collected data limit causal interpretation of individual changes.

\begin{figure}[t]
\centering
\includegraphics[width=0.9\linewidth]{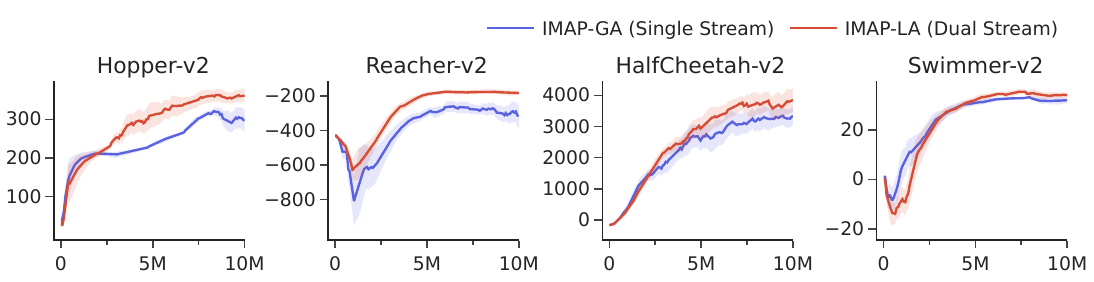}
\caption{Global and weighted local actor readouts on MAMuJoCo, with the same Q/V reward-model design and global critic. IMAP-LA has higher final means after 10M steps.}
\label{fig:mujoco_adv}
\end{figure}
On identical held-out pairs, the Q/V model has better reported ranking accuracy and negative log-likelihood (NLL) than the direct head on all three measured tasks (Table~\ref{tab:heldout_reward}). On Hopper, accuracy is $0.891$ versus $0.812$, and NLL is $0.517$ versus $0.634$. These three-seed descriptive means concern fitted team scores; they do not by themselves validate the local actor direction. The relay test addresses that direction.

SMACLite compares temporal-agent redistribution methods with cumulative native return revealed at termination \citep{michalski2023smaclitelightweightenvironmentmultiagent,xiao2022agent,chen2024stas,kapoor2025redistributingrewardstimeagents}. IMAP-LA has the highest mean on two scenarios and exceeds SparseMAPPO and AREL throughout; TAR$^2$ has the highest mean on 3s5z (Appendix Table~\ref{tab:smaclite_sota}).

\subsection{Which component errors change local credit?}\label{sec:toy}
We test the predicted error channels in a four-decision, two-agent terminal relay. At frozen checkpoints, Q-only and joint Q/V shifts preserve the mixed reward, policy and critic but change local credit through compatibility and critic--head continuation, respectively. At training $\lambda=0.95$ and unit amplitude, the mean normalized actor-direction responses across ten models are $0.11496$ and $0.00605$. Their opposite zero-response endpoints and all 440 checkpoint bound checks meet the declared numerical tolerances (Appendices~\ref{app:pathwise-gae}--\ref{app:numerics}; Appendix Figure~\ref{fig:toy_verification}). These tests diagnose frozen credit directions, not policy improvement; Appendix~\ref{app:signal-diagnostic} separately examines the structured signal result.

The Q-only response vanishes at $\lambda=0$, and the joint Q/V response
at $\lambda=1$. For the shared perturbations used here, their magnitudes
at training $\lambda$ differ by $\lambda/(1-\lambda)=19$, as predicted by
Equation~\eqref{eq:continuation-main}. Thus an unchanged team score can
support different local actor directions through the two continuation
channels.

\section{Conclusion}\label{sec:conclusion}
IMAP turns terminal outcomes into agent-specific online PPO updates. Comparisons between completed episodes fit shared Q/V heads; their residuals and normalized mixture give the centralized critic a team reward and decentralized actors weighted local advantages. With terminal-only feedback, IMAP-LA has the highest reported mean among the compared methods on nine SMACv2 and four MAMuJoCo tasks, and on two of three SMACLite scenarios. Direct-head and global-readout comparisons support the model and actor choices; frozen relay interventions confirm predicted credit responses at unchanged mixed reward.

An exact general-GAE identity and sharp full-return bound characterize local-credit error for the learned objective; a read-once theorem quantifies ideal task signal. Calibrating the fitted score to terminal outcomes and extending the analysis to finite-rollout clipped PPO and richer mixers are promising directions for future work.

\label{main-end}
\typeout{MAINPAPERENDPAGE:\thepage}
\clearpage

\subsection*{Statement of Use of AI}
The research questions, the main ideas, the methodology, and the experimental protocols of this paper were chosen by the authors, who divided the work among themselves, supervised every stage of the project, and made all decisions on claims and conclusions. Under the authors' direction, we used generative AI tools as assistants for the following tasks with required disclosure: refining the conceptual framing, stress-testing mathematical claims and proof arguments, and implementing and running the relay and numerical-verification experiments. In the theoretical component, generative AI was used in part as a “theory attacker”: it assisted in implementing computational searches and large-scale simulations over extensive collections of generated examples in order to identify potential counterexamples, edge cases, or weaknesses in candidate theorem statements and proof arguments. These searches were used to help the authors examine and refine the theoretical claims. The failure of such searches to identify a counterexample does not constitute a proof, and no theorem in this paper is considered established on the basis of AI-assisted simulation, search, or numerical testing.

The authors independently verified all theorem statements and proofs, citations, experimental specifications, and AI-assisted text against their research artifacts. No experimental data or results were produced by generative AI tools themselves: every reported number comes from experiments that were actually executed, and translation and qualitative data analysis are not applicable to this work. Additionally, we used generative AI tools for literature discovery and for drafting and editing the manuscript, including checks of notation and of the consistency of reported numbers. The authors take full responsibility for every claim, proof, citation, and experimental result in this paper; no AI system is an author.

\subsection*{Reproducibility statement}
Section~\ref{sec:setup} defines terminal feedback, and Section~\ref{sec:method} and Algorithm~\ref{alg:imap} specify IMAP. The local-credit and structured-signal proofs are in Appendices~\ref{app:theory} and~\ref{credit:signal-appendix}. Section~\ref{sec:experiments} reports benchmark results; Appendix~\ref{app:implementation} gives training and evaluation settings (Table~\ref{tab:benchmark_config}). The relay construction, checkpoint population, and numerical checks are specified in Appendices~\ref{app:toy-protocol}--\ref{app:numerics}; Appendix~\ref{app:signal-diagnostic} describes the fitted-signal diagnostic. The appendices describe the benchmark experiments, relay and numerical-verification studies, and fitted-signal diagnostics.

\bibliography{references_v1}
\bibliographystyle{iclr2027_conference}
\clearpage
\appendix
\section{Proofs of the local-credit results}\label{app:theory}

\subsection{Setting and quantities}\label{app:theory-setting}
Consider a finite-horizon Markov game with stages $h=0,\ldots,H-1$,
$H\ge1$, and a sufficient controlled information state $\mathbf{S}_h$ containing all actor
conditioning information. The joint history is one admissible choice. Agent $i$ observes $o_{i,h}(\mathbf{S}_h)$ and samples its action from
$\pi_{i,\theta}(\cdot\mid o_{i,h}(\mathbf{S}_h))$. Conditional on $\mathbf{S}_h$, actions are
independent throughout a neighborhood of the current parameter:
\[
\pi_\theta(\mathbf a\mid \mathbf{s})=\prod_{i=1}^n
\pi_{i,\theta}(a_i\mid o_{i,h}(\mathbf{s})).
\]
All agents may share the parameter vector $\theta\in\mathbb R^d$.
The initial distribution, transition kernels and observation maps are independent of $\theta$.
We first give the proof for finite state and action sets, with differentiable
policies positive on fixed feasible action sets in a neighborhood of the
current parameter. A forced no-op is a singleton action set and has zero
score. The extension to general state and action spaces is stated at the end.

Fix $\gamma,\lambda\in[0,1]$. Freeze real-valued local functions $q_{i,h}$,
$v_{i,h}$, a shared critic $V_h$, and scalar weights $\alpha_i\ge0$ with
$\sum_i\alpha_i=1$. Weights do not vary with the stage, state or action.
Each $q_{i,h}(o_i,a_i)$ depends on its own action only. Terminal values satisfy
$v_{i,H}=V_H=0$. All expectations and scores below are at the current policy,
and model quantities remain frozen when differentiating the learned return.
The reward and local residual are
\begin{align}
r_{i,h}&=q_{i,h}(o_{i,h},A_{i,h})-\gamma v_{i,h+1}(o_{i,h+1}),\label{pr:reward}\\
\widehat R_h&=\sum_i\alpha_i r_{i,h},\qquad
\delta_{i,h}=r_{i,h}+\gamma V_{h+1}(\mathbf{S}_{h+1})-V_h(\mathbf{S}_h).\label{pr:delta}
\end{align}
We suppress observation arguments when viewing local functions as functions
of $\mathbf{s}$. Define complete-episode advantages and learned returns by
\[
\widehat A^L_{i,h}=\sum_{t=h}^{H-1}(\gamma\lambda)^{t-h}\delta_{i,t},\qquad
\widehat G_h=\sum_{t=h}^{H-1}\gamma^{t-h}\widehat R_t,\qquad
J_{\widehat R}(\theta)=\mathbb E_\theta\widehat G_0.
\]
Let $\widehat A^G_h$ use the same GAE sum with residual
$\widehat R_h+\gamma V_{h+1}-V_h$. Normalization of the mixer gives the
pathwise identity
\begin{equation}
\widehat A^G_h=\sum_i\alpha_i\widehat A^L_{i,h}.
\label{pr:gae}
\end{equation}
Indeed, mixing each local residual gives the global residual, and the
constant weights commute with the common finite GAE sum.

For general GAE, use the unnormalized discounted time weights in
\[
z_{i,h}=\nabla_\theta\log\pi_{i,\theta}(A_{i,h}\mid o_{i,h}(\mathbf{S}_h)),\qquad
g_L^{(\lambda)}=\mathbb E\sum_{h=0}^{H-1}\sum_i\gamma^h\alpha_i z_{i,h}\widehat A^L_{i,h}.
\]
The derivative acts on the policy only. 

Write
\[
e_{i,h}=\mathbb E_{a_i\sim\pi_i}q_{i,h}(o_i,a_i)-v_{i,h},\qquad
D_{i,h}=V_h-v_{i,h},\qquad
\bar e_h=\sum_i\alpha_i e_{i,h},\qquad
\bar v_h=\sum_i\alpha_i v_{i,h}.
\]
Let $P_h^{\mathbf a}$ be the next-state kernel conditional on the joint action,
$P_{i,h}^{a_i}$ average the other agents' actions, and $P_{\pi,h}$ average all
actions under the current policy. For a state function $f$, $(Pf)(\mathbf{s})$ denotes
its expectation under the relevant next-state kernel. Define the finite
continuation operator
\begin{equation}
(\mathcal R_c f)_h(\mathbf{s})=\sum_{\ell=0}^{H-h-1}c^\ell
\mathbb E[f_{h+\ell}(\mathbf{S}_{h+\ell})\mid \mathbf{S}_h=\mathbf{s}],\qquad
(\mathcal R_c f)_H=0.
\label{pr:resolvent}
\end{equation}
An empty sum is zero, and a zeroth power is one, including when its base is
zero. Thus $\mathcal R_c f$ is the unique backward solution of
$u_h=f_h+cP_{\pi,h}u_{h+1}$ with $u_H=0$.

\subsection{Consistency of the global and local actor coefficients}
\label{app:objective-consistency}

The coefficient $\alpha_i$ in the local actor follows from the same
mixture that defines the learned team objective. The following calculation
records this relationship; it does not assert that the two stochastic
updates or their finite PPO trajectories coincide.
Use the assumptions and frozen quantities of
Appendix~\ref{app:theory-setting}, and define the unit-prefactor global vector
\begin{equation}
g_G^{(\lambda)}=
\mathbb E\sum_{h=0}^{H-1}\sum_i\gamma^h
z_{i,h}\widehat A^G_h.
\label{oc:global-vector}
\end{equation}
The expectations are at the old policy. The advantages are fixed targets,
without a nonlinear or sample-dependent advantage transformation.

\paragraph{One-stage equality.}
For $H=1$, every $\lambda\in[0,1]$ gives
\begin{equation}
g_G^{(\lambda)}=g_L^{(\lambda)}
=\nabla J_{\widehat R}
=\sum_i\alpha_i\mathbb E[z_{i,0}q_{i,0}].
\label{oc:one-stage}
\end{equation}
In contrast, applying an additional actor prefactor $\alpha_i$ to the
global advantage gives
\begin{equation}
\widetilde g_G
:=\mathbb E\sum_i\alpha_i z_{i,0}\widehat A^G_0
=\sum_i\alpha_i^2\mathbb E[z_{i,0}q_{i,0}],
\label{oc:extra-weight}
\end{equation}
which generally changes the reward gradient.

\begin{proof}
With zero terminal values, $r_{i,0}=q_{i,0}$ and
\[
\widehat A^L_{i,0}=q_{i,0}-V_0(\mathbf{S}_0),\qquad
\widehat A^G_0=\sum_j\alpha_jq_{j,0}-V_0(\mathbf{S}_0).
\]
Policy normalization implies
$\mathbb E[z_{i,0}\mid \mathbf{S}_0]=0$.
For $j\ne i$, conditional action independence and the own-action
restriction on $q_j$ give
\[
\mathbb E[z_{i,0}q_{j,0}\mid \mathbf{S}_0]
=\mathbb E[z_{i,0}\mid \mathbf{S}_0]\,
  \mathbb E[q_{j,0}\mid \mathbf{S}_0]=0.
\]
The shared critic is also a baseline, regardless of its accuracy.
Substitution into the two actor vectors leaves precisely the right-hand
side of Equation~\eqref{oc:one-stage}. Since the initial law and frozen
heads do not depend on actor parameters, likelihood-ratio differentiation
of $J_{\widehat R}=\mathbb E\sum_j\alpha_jq_{j,0}$ gives the same expression.
Inserting the extra prefactor in the global vector gives
Equation~\eqref{oc:extra-weight}. These conditional cancellations hold
coordinatewise even when actor parameters are shared.
\end{proof}

\paragraph{Full-return sequential extension.}
For $\lambda=1$ on a complete episode,
$\widehat A^G_h=\widehat G_h-V_h(\mathbf{S}_h)$, where
$\widehat G_h=\sum_{t=h}^{H-1}\gamma^{t-h}\widehat R_t$.
The policy-gradient identity and baseline cancellation therefore give
$g_G^{(1)}=\nabla J_{\widehat R}$ for any frozen critic.
Theorem~\ref{sf:theorem} and its stagewise refinement in
Appendix~\ref{app:sharp-proof} yield
\begin{equation}
\|g_L^{(1)}-g_G^{(1)}\|_{\mathcal I_\pi^\dagger}
\le\frac12\sum_{t=1}^{H-1}\gamma^t
\max_j\operatorname{span}e_{j,t}.
\label{oc:sequential}
\end{equation}
In particular, the two expected reward gradients agree when every
$e_{j,t}$ is constant across stage-$t$ states for $t\ge1$; those constants
need not be zero or equal across components. The discounted time weights
$\gamma^h$ in both vectors are those of
$J_{\widehat R}=\mathbb E\sum_h\gamma^h\widehat R_h$.

\paragraph{Entropy and implementation scope.}
Both actor objectives use the same entropy coefficient outside the reward
prefactor. At a common old policy and with the same state-sampling measure
and reduction, their entropy derivatives are identical. Hence, whenever
the expected reward vectors above agree, adding the same entropy term
preserves that equality. This observation does not compare the random
reward-gradient magnitudes or their subsequent balance with entropy.

The statements concern expected derivatives at the old policy, where
positive-width PPO clipping is locally inactive. They do not establish
equal minibatch gradients, variance reduction, or equality after finite
steps and repeated clipped updates. For $H>1$ and $\lambda<1$, the
continuation terms in Appendix~\ref{app:general-gae} must be retained;
Equation~\eqref{oc:sequential} does not apply directly to the training
setting $\lambda=0.95$. Rollout truncation requires the endpoint terms of
Appendix~\ref{app:bootstrap}. A fixed common rescaling of both reward
vectors rescales these equalities and bounds, whereas empirical
agent-specific or sample-dependent advantage normalization and different
time reductions require separate analysis. Finally, an additional-weight
global variant remains a useful empirical control, but
Equation~\eqref{oc:extra-weight} distinguishes it from the global gradient
of the original mixed objective.

\subsection{Proof of Theorem~\ref{sf:theorem}}\label{app:sharp-proof}
\begin{proof}
Write $L_H=\sum_{t=1}^{H-1}\gamma^t$, with $L_1=0$.
Let $G_j=\sum_{t=0}^{H-1}\gamma^t r_{j,t}$, $S_j=\sum_hz_{j,h}$ and
$S_{-j}=\sum_{h,i\ne j}z_{i,h}$.
At $\lambda=1$ and $V_H=0$, the shared-critic terms telescope:
\[
\widehat A^L_{i,h}=\sum_{t=h}^{H-1}\gamma^{t-h}r_{i,t}-V_h(\mathbf{S}_h).
\]
Policy normalization gives zero conditional mean for $z_{i,h}$ given the
pre-action history. Hence its product with $V_h(\mathbf{S}_h)$ has zero expectation.
For $t<h$, $r_{i,t}$ is measurable before the stage-$h$ action, so
$\mathbb E[z_{i,h}r_{i,t}]=0$. We may therefore replace the discounted
return-to-go in the actor sum by the complete local return. Likelihood-ratio
differentiation of the finite trajectory law gives the team-return gradient:
\[
g_L^{(1)}=\sum_j\alpha_j\mathbb E[S_jG_j],\qquad
\nabla J_{\widehat R}=\sum_j\alpha_j\mathbb E[SG_j],
\]
and their difference is $-\sum_j\alpha_j\mathbb E[S_{-j}G_j]$.
This cancellation uses a frozen critic; its accuracy is not required.

The Q/V reward telescopes pathwise:
\[
G_j=v_{j,0}(\mathbf{S}_0)+\sum_{t=0}^{H-1}\gamma^t
\big(q_{j,t}(o_{j,t},A_{j,t})-v_{j,t}(o_{j,t})\big).
\]
To compute its product with a score $z_{i,h}$ for $i\ne j$, consider three
cases. If $h<t$, condition on the full pre-action history at $t$: the score
is already measurable and the bracket averages to $e_{j,t}(\mathbf{S}_t)$.
If $h=t$, conditional action independence and zero mean of the own-agent
score make the product zero; the product with $e_{j,t}(\mathbf{S}_t)$ is also zero.
If $h>t$, the bracket is measurable before the later action and the later
score has conditional mean zero. The same holds for $e_{j,t}(\mathbf{S}_t)$.
The initial value and $t=0$ compatibility terms also have zero product with
$S_{-j}$. Consequently
\begin{equation}
b:=g_L^{(1)}-\nabla J_{\widehat R}
=-\sum_j\alpha_j\mathbb E[S_{-j}T_j],\qquad
T_j=\sum_{t=1}^{H-1}\gamma^t e_{j,t}(\mathbf{S}_t).
\label{sf:cross}
\end{equation}
Here $e_j$ is the function evaluated at the current policy and then fixed.
No differentiation through its definition is taken in Equation~\eqref{sf:cross}.

Distinct score increments are orthogonal in second moment. Across stages
this follows by conditioning on the later pre-action history. Within a
stage it follows from conditional action independence and zero conditional
means. Therefore, even with shared parameter coordinates,
\begin{equation}
\mathbb E[S_{-j}S_{-j}^\top]
=\sum_{h,i\ne j}\mathbb E[z_{i,h}z_{i,h}^\top]
\preceq\sum_{h,i}\mathbb E[z_{i,h}z_{i,h}^\top]
=\mathcal I_\pi.
\label{sf:score-order}
\end{equation}

The range of $T_j$ has length at most $L_H\varepsilon$. A random variable
in an interval of length $r$ has variance at most $r^2/4$: subtract its
midpoint and use that the mean minimizes squared error over constants.
Since $\mathbb E S_{-j}=0$, Cauchy--Schwarz gives, for every vector $u$,
\begin{align*}
|u^\top b|
&\le\sum_j\alpha_j
 \sqrt{\mathbb E[(u^\top S_{-j})^2]\operatorname{Var}(T_j)}\\
&\le\sqrt{u^\top\mathcal I_\pi u}\,\frac{L_H\varepsilon}{2}.
\end{align*}
In particular, every vector in the kernel of $\mathcal I_\pi$ is orthogonal
to $b$, proving the range assertion. Taking the supremum over
$u^\top\mathcal I_\pi u\le1$ proves Equation~\eqref{sf:bound}.
\end{proof}

\paragraph{Exact agreement and effective duration.}
If each compatibility error is constant across states at each stage,
$\varepsilon=0$ and the two expected gradients agree. More finely, the same
proof gives
\[
\|b\|_{\mathcal I_\pi^\dagger}
\le\frac12\sum_{t=1}^{H-1}\gamma^t
\max_j\operatorname{span} e_{j,t}.
\]
Only stages with nonconstant compatibility error contribute to this
certificate. An inert suffix with state-constant errors adds no term.
There is no factor depending on the agent count, parameter dimension or a
transition-contraction coefficient. This describes the total expected actor
vector, including all stages and shared parameters.

\subsection{Sharpness of the constant}\label{app:sharpness}
Fix $H\ge2$, $\gamma>0$ and $\varepsilon>0$. Agent 1 makes a single binary
decision at stage zero, with probability $p=\operatorname{sigmoid}(\theta)$
of the positive branch. The next state records the branch, and all remaining
actions of both agents are forced no-ops. Agent 2 observes the branch.
Take all values and the shared critic to be zero, $q_1=0$, $q_{2,0}=0$, and
at every remaining stage set
\[
q_{2,t}=e_{2,t}=
\begin{cases}\varepsilon/2,&\text{positive branch},\\
-\varepsilon/2,&\text{negative branch}.
\end{cases}
\]
With $\alpha=(0,1)$, the local actor vector is zero. The team return is
$J_{\widehat R}=L_H\varepsilon(p-1/2)$. At $p=1/2$,
\[
|\nabla J_{\widehat R}|=L_H\varepsilon/4,\qquad
\mathcal I_\pi=p(1-p)=1/4,
\]
so Equation~\eqref{sf:bound} is an equality. With the strictly positive
mixer $(\eta,1-\eta)$, the left side is
$(1-\eta)L_H\varepsilon/2$, approaching the bound as $\eta\downarrow0$.
Thus no smaller constant works uniformly over strictly positive normalized
mixers either. Sharpness is uniform over mixers, not attainment for every
fixed positive mixer.

The environmental reward can be restricted to the terminal binary branch
outcome. The learned return is a positive affine transform of that outcome
when $\gamma>0$.

\subsection{Pathwise local GAE and rollout boundaries}
\label{app:pathwise-gae}

For a fixed sampled rollout from stage $h$ to its boundary $H$, write
$D_{i,t}=V_t-v_{i,t}$ and $c=\gamma\lambda$.
The endpoint values are the declared frozen bootstrap values, which can
be nonzero at a rollout cutoff. Define finite GAE with the same rewards,
critic and endpoint masks as in Equation~\eqref{eq:gae}. Then
\begin{equation}
\begin{aligned}
\widehat A^L_{i,h}
={}&\sum_{t=h}^{H-1}c^{t-h}(q_{i,t}-v_{i,t})-D_{i,h}\\
&+\gamma(1-\lambda)\sum_{t=h}^{H-2}c^{t-h}D_{i,t+1}
+\gamma c^{H-h-1}D_{i,H}.
\end{aligned}
\label{eq:local-gap-gae-boundary}
\end{equation}
All identities are samplewise; no expectation, action-factorization or
calibration assumption is needed. Empty sums equal zero, and powers with
exponent zero equal one, including at $c=0$. At a true terminal boundary,
$v_{i,H}=V_H=0$, giving
\begin{equation}
\begin{aligned}
\widehat A^L_{i,h}
={}&\sum_{t=h}^{H-1}(\gamma\lambda)^{t-h}
       (q_{i,t}-v_{i,t})-D_{i,h}\\
&+\gamma(1-\lambda)\sum_{t=h}^{H-2}(\gamma\lambda)^{t-h}
       D_{i,t+1}.
\end{aligned}
\label{eq:local-gap-gae}
\end{equation}

\begin{proof}
The implicit reward and shared critic residual give
\[
\delta^L_{i,t}
=q_{i,t}-\gamma v_{i,t+1}+\gamma V_{t+1}-V_t
=(q_{i,t}-v_{i,t})-D_{i,t}+\gamma D_{i,t+1}.
\]
Multiply by $c^{t-h}$ and sum from $t=h$ through $H-1$.
The coefficient of $D_{i,h}$ is $-1$. For any interior index
$u\in\{h+1,\ldots,H-1\}$, its coefficient is
\[
\gamma c^{u-h-1}-c^{u-h}
=\gamma(1-\lambda)c^{u-h-1}.
\]
The endpoint coefficient is $\gamma c^{H-h-1}$.
Collecting these terms proves Equation~\eqref{eq:local-gap-gae-boundary},
including $\lambda=0$ and $\gamma=0$, without division by $c$.
\end{proof}

For a trajectory that terminates before a batch rollout ends, apply the
identity up to its true terminal boundary, or equivalently use an absorbing
zero-valued suffix and the same terminal masks throughout. A truncation
with bootstrapping retains the displayed endpoint term; treating it as a
true terminal transition would change the readout. The population-gradient
correction for the declared bootstrapped objective is given separately in
Appendix~\ref{app:bootstrap}.

At $\lambda=1$ the interior $D$ terms vanish, so the difference from the
discounted sum of local Q/V gaps consists only of the current baseline and,
if present, the boundary term. At $\lambda<1$, the coefficient
$(1-\lambda)$ is exact, but the full accumulated correction need not be
small without a bound on future $D$ and the effective rollout length.
The current-state term has zero expected old-policy score product;
it can still affect finite-sample variance and later clipped PPO updates.
Thus the identity does not assert automatic variance reduction.

\paragraph{Fixed-score responses.}
Fix locally measurable state potentials $\phi_{i,h}$ with
$\sum_i\alpha_i\phi_{i,h}=0$ pointwise and $\phi_{i,H}=0$.
They and the mixer are frozen during actor differentiation.
Under a joint Q/V shift of amplitude $\zeta$, the local Q/V gap is
unchanged and $D_i$ changes to $D_i-\zeta\phi_i$.
Under a Q-only shift, the local Q/V gap changes by $\zeta\phi_i$
and $D_i$ is unchanged. The mixed Q and V, hence mixed rewards and
global critic targets, are unchanged in both cases.

Set $c=\gamma\lambda$ and define
\[
K_\phi=
\E\sum_{h,i}\gamma^h\alpha_i z_{i,h}
\sum_{t=h}^{H-2}c^{t-h}\phi_{i,t+1}(\mathbf{S}_{t+1}).
\]
The current-state potential has zero expected score product:
$\E[z_{i,h}\phi_{i,h}(\mathbf{S}_h)]=0$ by conditioning on the pre-action
state. Applying Equation~\eqref{eq:local-gap-gae} to the joint shift
therefore gives
\[
\Delta g_{QV}=-\zeta\gamma(1-\lambda)K_\phi.
\]
For the Q-only shift, direct summation of its GAE change gives
$\zeta\sum_{t=h}^{H-1}c^{t-h}\phi_{i,t}(\mathbf{S}_t)$.
Its current-state term cancels in expectation; reindexing the remainder
gives
\[
\Delta g_Q=\zeta\gamma\lambda K_\phi.
\]
Combining the response formulas yields
\begin{equation}
(1-\lambda)\Delta g_Q+\lambda\Delta g_{QV}=0.
\label{eq:gae-channel-response}
\end{equation}
The relation includes
the endpoints $\lambda=0,1$ without division. They permit shared actor
parameters. The response identity also holds with a common frozen
pre-action-measurable time weight in place of $\gamma^h$, since the
same current-state cancellation and reindexing apply within each
stage's advantage. It concerns expected reward-gradients at the old
policy, before a nonlinear advantage transform or subsequent clipped
PPO updates. A common fixed scalar normalization preserves the relation.

For the Fisher responses in Figure~\ref{fig:toy_verification}, the policy,
Fisher matrix, potential amplitude and original-head scale are identical
between the two interventions. Homogeneity of the norm thus gives
$\|\Delta g_Q\|=19\|\Delta g_{QV}\|$ at $\lambda=0.95$, with the
corresponding endpoint zeros. The controlled ratio describes sensitivity
at fixed mixed reward; it is not a comparison of policy returns.

\subsection{Exact continuation identity for general GAE}\label{app:general-gae}
\begin{lemma}[Continuation discrepancy]\label{pr:identity-thm}
Define
\begin{equation}
F_i=\alpha_i\mathcal R_{\gamma\lambda}
\big((1-\lambda)D_i+\lambda e_i\big)-\mathcal R_\gamma\bar e.
\label{pr:F}
\end{equation}
Under the preceding assumptions,
\begin{equation}
\boxed{
g_L^{(\lambda)}-\nabla J_{\widehat R}
=\mathbb E\sum_{h=0}^{H-1}\sum_i\gamma^{h+1}z_{i,h}
\big[(P_{i,h}^{A_{i,h}}-P_{\pi,h})F_{i,h+1}\big](\mathbf{S}_h).}
\label{pr:identity}
\end{equation}
This holds with shared policy parameters and with zero mixer weights.
\end{lemma}
\begin{proof}
First establish the learned-return policy gradient. Differentiating the
finite trajectory sum, only policy factors depend on $\theta$, so
\[
\nabla J_{\widehat R}
=\mathbb E\left[\widehat G_0\sum_{h,i}z_{i,h}\right].
\]
Let $\mathcal H_h$ be the full pre-action history. Policy normalization and
conditioning only on $\mathbf{S}_h$ give
$\mathbb E[z_{i,h}\mid\mathcal H_h]=\mathbb E[z_{i,h}\mid \mathbf{S}_h]=0$.
Rewards preceding stage $h$ are $\mathcal H_h$-measurable, so their products
with that stage's score have zero expectation. The remaining terms are
\begin{equation}
\nabla J_{\widehat R}
=\mathbb E\sum_{h,i}\gamma^h z_{i,h}\widehat G_h.
\label{pr:pg}
\end{equation}
This derivation sums each agent's score even when the scores use the same
parameter coordinates.

We next compute the two conditional continuation values. Let
$M_{i,h}(\mathbf{s})=\mathbb E[\widehat A^L_{i,h}\mid \mathbf{S}_h=\mathbf{s}]$ and $M_{i,H}=0$.
Equation~\eqref{pr:delta} is
$\delta_{i,h}=q_{i,h}-V_h+\gamma D_{i,h+1}$. Conditioning the GAE recursion
on the current state gives
\[
M_{i,h}=e_{i,h}-D_{i,h}
+\gamma P_{\pi,h}D_{i,h+1}
+\gamma\lambda P_{\pi,h}M_{i,h+1}.
\]
Set $W_{i,h}=D_{i,h}+\lambda M_{i,h}$. Multiplying the preceding recursion
by $\lambda$ and adding $D_{i,h}$ gives
\[
W_{i,h}=(1-\lambda)D_{i,h}+\lambda e_{i,h}
+\gamma\lambda P_{\pi,h}W_{i,h+1},\qquad W_{i,H}=0.
\]
There is no division by $\lambda$. By backward substitution,
\begin{equation}
W_i=\mathcal R_{\gamma\lambda}
\big((1-\lambda)D_i+\lambda e_i\big).
\label{pr:W}
\end{equation}
Consequently, conditional on the current joint action,
\begin{equation}
\mathbb E[\widehat A^L_{i,h}\mid \mathbf{s},\mathbf a]
=q_{i,h}(o_i,a_i)-V_h(\mathbf{s})+\gamma(P_h^{\mathbf a}W_{i,h+1})(\mathbf{s}).
\label{pr:conditional-local}
\end{equation}

For the team return, let $U_h(\mathbf{s})=\mathbb E[\widehat G_h\mid \mathbf{S}_h=\mathbf{s}]$.
Its Bellman recursion is
\[
U_h=\bar v_h+\bar e_h
+\gamma P_{\pi,h}(U_{h+1}-\bar v_{h+1}),\qquad U_H=\bar v_H=0.
\]
Thus $U-\bar v=\mathcal R_\gamma\bar e$, and, writing
$\bar q_h(\mathbf{s},\mathbf a)=\sum_j\alpha_jq_{j,h}(o_j,a_j)$,
\begin{equation}
\mathbb E[\widehat G_h\mid \mathbf{s},\mathbf a]
=\bar q_h(\mathbf{s},\mathbf a)
+\gamma\big(P_h^{\mathbf a}(\mathcal R_\gamma\bar e)_{h+1}\big)(\mathbf{s}).
\label{pr:conditional-team}
\end{equation}

Subtract Equation~\eqref{pr:conditional-team} from $\alpha_i$ times
Equation~\eqref{pr:conditional-local}. The result is
\begin{equation}
-\sum_{j\ne i}\alpha_j q_{j,h}(o_j,a_j)-\alpha_i V_h(\mathbf{s})
+\gamma(P_h^{\mathbf a}F_{i,h+1})(\mathbf{s}).
\label{pr:pre-cancel}
\end{equation}
The shared-critic term is independent of $a_i$. For $j\ne i$,
conditional factorization and the own-action dependence of $q_j$ give
\[
\mathbb E[z_{i,h}q_{j,h}(o_j,A_{j,h})\mid \mathbf{S}_h=\mathbf{s}]
=\mathbb E[z_{i,h}\mid \mathbf{S}_h=\mathbf{s}]\,
 \mathbb E[q_{j,h}(o_j,A_{j,h})\mid \mathbf{S}_h=\mathbf{s}]=0.
\]
This cancellation remains valid with shared parameters: it uses conditional
action independence, not disjoint parameter coordinates. Combining
Equation~\eqref{pr:pg} with Equation~\eqref{pr:pre-cancel} therefore yields
\[
g_L^{(\lambda)}-\nabla J_{\widehat R}
=\mathbb E\sum_{h,i}\gamma^{h+1}z_{i,h}
(P_h^{\mathbf A_h}F_{i,h+1})(\mathbf{S}_h).
\]
Averaging other agents' actions replaces $P_h^{\mathbf A_h}$ by
$P_{i,h}^{A_{i,h}}$. Finally,
conditional on $\mathbf{S}_h$, $(P_{\pi,h}F_{i,h+1})(\mathbf{S}_h)$ does not depend on the
sampled $A_{i,h}$, so its product with $z_{i,h}$ has zero conditional mean. Subtracting it proves
Equation~\eqref{pr:identity}.
\end{proof}

\paragraph{The retained GAE setting.}
The reference training value $\lambda=0.95$ gives
\begin{equation}
F_i^{(0.95)}=\alpha_i\mathcal R_{0.95\gamma}
(0.05D_i+0.95e_i)-\mathcal R_\gamma\bar e.
\label{pr:training-continuation}
\end{equation}
Substitution in Equation~\eqref{pr:identity} gives the exact discrepancy of
$g_L^{(0.95)}$ under the stated time weights and episode boundary.
This is the continuation used when evaluating the training readout.
At $\lambda=1$, the same formula reduces to
$F_i=-\mathcal R_\gamma\sum_{j\ne i}\alpha_j e_j$, which underlies
Equation~\eqref{sf:cross}.

\subsection{Finite-rollout endpoints}\label{app:bootstrap}
Fix a rollout boundary $H$ and a frozen terminal target $b_H(\mathbf{S}_H)$.
The comparison objective is the bootstrapped return
$J_{\widehat R,b}=\mathbb E[\sum_{t=0}^{H-1}\gamma^t\widehat R_t+\gamma^H b_H(\mathbf{S}_H)]$.
The local GAE still sums the $H$ observed residuals, with the declared endpoint
values $v_{i,H}$ and $V_H$. Let
$(P_{\pi,h:H}f)(\mathbf{s})=\mathbb E[f(\mathbf{S}_H)\mid \mathbf{S}_h=\mathbf{s}]$ and
$\bar v_H=\sum_j\alpha_jv_{j,H}$. Replace $F_i$ in Lemma~\ref{pr:identity-thm} by
\begin{align}
\widetilde F_{i,h}=F_{i,h}
&+\alpha_i(\gamma\lambda)^{H-h}P_{\pi,h:H}(V_H-v_{i,H})\nonumber\\
&+\gamma^{H-h}P_{\pi,h:H}(\bar v_H-b_H).
\label{pr:boundary}
\end{align}
Then Equation~\eqref{pr:identity} holds with $\widetilde F_i$ and
$\nabla J_{\widehat R,b}$. To verify it, the local recursion now has
$W_{i,H}=V_H-v_{i,H}$, and the team recursion has
$(U-\bar v)_H=b_H-\bar v_H$. Backward substitution adds exactly the two
terms in Equation~\eqref{pr:boundary}; the same score cancellation completes
the identity. For a critic bootstrap use $b_H=V_H$; for the truncated reward
sum use $b_H=0$. Zero terminal values recover the complete-episode result.
The bootstrap objective is explicitly the declared frozen-endpoint target.

\subsection{Technical bounds for general GAE}\label{app:general-bounds}
Throughout this subsection, $\|\cdot\|$ denotes the Euclidean norm.
For a real-valued function on the stage-$h$ state set, write
$\operatorname{span}f=\max_{\mathbf{s}}f(\mathbf{s})-\min_{\mathbf{s}}f(\mathbf{s})$, and let
$\operatorname{TV}(p,q)=\tfrac12\sum_{\mathbf{s}}|p(\mathbf{s})-q(\mathbf{s})|$.
Define
\begin{align}
\eta_{i,h}(\mathbf{s},a_i)&=\operatorname{TV}
  (P_{i,h}^{a_i}(\cdot\mid \mathbf{s}),P_{\pi,h}(\cdot\mid \mathbf{s})),\\
\mathcal C_\pi&=\mathbb E\sum_{h,i}\gamma^{h+1}
  \|z_{i,h}\|\eta_{i,h}(\mathbf{S}_h,A_{i,h}).\label{pr:C}
\end{align}
For any probability vectors $p,q$ and function $f$,
\begin{equation}
|(p-q)f|\le\operatorname{TV}(p,q)\operatorname{span}f.
\label{pr:tv-span}
\end{equation}
To see the constant exactly, subtract the midpoint of the range of $f$.
The sum of absolute signed masses is $2\operatorname{TV}(p,q)$ and the
centered function has supremum norm at most $\operatorname{span}f/2$.

\begin{corollary}[Action influence and continuation variation]\label{pr:compact-thm}
The continuation discrepancy satisfies
\begin{align}
\|g_L^{(\lambda)}-\nabla J_{\widehat R}\|
&\le\mathbb E\sum_{h,i}\gamma^{h+1}\|z_{i,h}\|
  \eta_{i,h}(\mathbf{S}_h,A_{i,h})\operatorname{span}F_{i,h+1}\label{pr:stage-bound}\\
&\le\mathcal C_\pi\max_{i,\,1\le h\le H}\operatorname{span}F_{i,h}.
\label{pr:compact}
\end{align}
\end{corollary}
\begin{proof}
Apply the triangle inequality to Equation~\eqref{pr:identity}, then apply
Equation~\eqref{pr:tv-span} inside the expectation. All coefficients are
nonnegative. Taking out the maximum continuation span gives the second
inequality. The terminal function $F_{i,H}$ is zero, so the maximum is
well-defined also at $H=1$.
\end{proof}

\paragraph{Technical refinement in local Q/V quantities.}
The compact bound can be evaluated through stagewise compatibility and value
variation. Let $\kappa\in[0,1]$ satisfy, at every stage,
\[
\sup_{\mathbf{s},\mathbf{s}'}\operatorname{TV}(P_{\pi,h}(\cdot\mid \mathbf{s}),
P_{\pi,h}(\cdot\mid \mathbf{s}'))\le\kappa.
\]
Set
\[
\varepsilon_Q=\max_{h<H,i}\operatorname{span}e_{i,h},\qquad
\varepsilon_V=\max_{h<H,i}\operatorname{span}D_{i,h},\qquad
K_c(H)=\sum_{\ell=0}^{H-2}c^\ell,
\]
where $K_c(1)=0$.
\begin{lemma}[Compatibility and GAE bound]\label{pr:expanded-thm}
Under the preceding assumptions,
\begin{equation}
\begin{aligned}
\|g_L^{(\lambda)}-\nabla J_{\widehat R}\|
\le\mathcal C_\pi\big[&K_{\gamma\lambda\kappa}(H)
 ((1-\lambda)\varepsilon_V+\lambda\varepsilon_Q)\\
&+K_{\gamma\kappa}(H)\varepsilon_Q\big].
\end{aligned}
\label{pr:expanded}
\end{equation}
\end{lemma}
\begin{proof}
Equation~\eqref{pr:tv-span}, applied to two rows of $P_{\pi,h}$, implies
$\operatorname{span}(P_{\pi,h}f)\le\kappa\operatorname{span}f$.
A product of $\ell$ such kernels contracts spans by at most $\kappa^\ell$,
by repeated application. Hence, for $c\ge0$ and $h<H$,
\begin{equation}
\operatorname{span}(\mathcal R_c f)_h
\le\sum_{\ell=0}^{H-h-1}(c\kappa)^\ell
     \operatorname{span}f_{h+\ell}.
\label{pr:resolvent-span}
\end{equation}
Nonnegative normalized weights give
$\operatorname{span}\bar e_h\le\sum_j\alpha_j
\operatorname{span}e_{j,h}\le\varepsilon_Q$. Likewise,
\[
\operatorname{span}\big((1-\lambda)D_{i,h}+\lambda e_{i,h}\big)
\le(1-\lambda)\varepsilon_V+\lambda\varepsilon_Q.
\]
Using Equation~\eqref{pr:F} and Equation~\eqref{pr:resolvent-span}, for
$1\le h<H$ we obtain
\begin{align*}
\operatorname{span}F_{i,h}
\le{}&\alpha_i\sum_{\ell=0}^{H-h-1}(\gamma\lambda\kappa)^\ell
  ((1-\lambda)\varepsilon_V+\lambda\varepsilon_Q)\\
&+\sum_{\ell=0}^{H-h-1}(\gamma\kappa)^\ell\varepsilon_Q.
\end{align*}
Because $\alpha_i\le1$ and $h\ge1$, each sum is at most its corresponding
$K_c(H)$. The terminal span is zero. Substitution in
Equation~\eqref{pr:compact} proves the result. No strict contraction is
required: $\kappa=1$ is allowed because all sums are finite.
\end{proof}

\paragraph{A sharper quantity for exact diagnostics.}
The proof also gives a computable stage- and agent-specific certificate.
With $b_{i,h}=(1-\lambda)D_{i,h}+\lambda e_{i,h}$, set
\[
B_{i,h}=\alpha_i\sum_{\ell=0}^{H-h-1}(\gamma\lambda\kappa)^\ell
\operatorname{span}b_{i,h+\ell}
+\sum_{\ell=0}^{H-h-1}(\gamma\kappa)^\ell
\operatorname{span}\bar e_{h+\ell},\qquad B_{i,H}=0.
\]
Then $\operatorname{span}F_{i,h}\le B_{i,h}$. Substituting $B_{i,h+1}$ in
Equation~\eqref{pr:stage-bound} retains the actual mixer weights and remaining
horizon. For each parameter coordinate $k$, the same argument replaces
$\|z_{i,h}\|$ by $|z_{i,h,k}|$ and bounds
$|(g_L^{(\lambda)}-\nabla J_{\widehat R})_k|$. These certificates follow from the same
proof and do not require extra model-fitting assumptions.

\begin{corollary}[Exact agreement]\label{pr:alignment}
If
$[(P_{i,h}^{a_i}-P_{\pi,h})F_{i,h+1}](\mathbf{s})=0$ for every reached
$(h,\mathbf{s},i,a_i)$ with positive policy probability, then
$g_L^{(\lambda)}=\nabla J_{\widehat R}$. In particular, agreement holds if
$\lambda=1$ and every $e_{i,h}$ is constant across stage-$h$ states.
It also holds when all actions leave the next-state distribution unchanged,
when $H=1$, or when $\gamma=0$.
\end{corollary}
\begin{proof}
The first condition sets every summand in Equation~\eqref{pr:identity}
to zero. At $\lambda=1$ and $\varepsilon_Q=0$,
Equation~\eqref{pr:expanded} is zero, whatever $\varepsilon_V$.
If $P_{i,h}^{a_i}=P_{\pi,h}$ for all reached actions, the transition contrasts
vanish. When $H=1$, only $F_{i,1}=0$ enters; when $\gamma=0$, every
$\gamma^{h+1}$ is zero.
\end{proof}
At $\lambda=1$, Equation~\eqref{pr:F} simplifies further to
\[
F_i=-\mathcal R_\gamma\sum_{j\ne i}\alpha_j e_j.
\]
Thus the acting agent's own compatibility term cancels from its continuation
discrepancy; the remaining terms come from the other components. In
particular, a one-agent model has exact agreement at $\lambda=1$ for arbitrary
local compatibility error.
The first condition is sufficient. The sum can also vanish through score
projection or cancellation between its terms. The value discrepancies
$e_i$ and $D_i$ therefore need not vanish to obtain exact expected-gradient
agreement.

\subsection{Technical transfer to terminal return}\label{app:terminal-transfer}
The following lemma uses a Euclidean credit-error certificate, such as
Equation~\eqref{pr:compact} or Equation~\eqref{pr:expanded}, at the chosen
$\lambda$.
Let $J^*(\vartheta)=\mathbb E_{\pi_\vartheta}G^*$ denote the task's expected
team return. Fix $a>0$ and $b\in\mathbb R$ for the frozen reward model.
For the current parameter $\theta$ and the actual updated parameter
$\theta+u$, suppose
\begin{equation}
|J_{\widehat R}(\vartheta)-aJ^*(\vartheta)-b|
\le\varepsilon_{\mathrm{out}},\qquad
\vartheta\in\{\theta,\theta+u\}.
\label{pr:calibration}
\end{equation}
A uniform condition on any class containing these two policies implies this
endpoint condition. Neither the intermediate segment nor unrelated policies
need outcome calibration for the result below.

\begin{lemma}[Terminal-gain certificate]\label{pr:gain-thm}
Assume the endpoint condition in Equation~\eqref{pr:calibration}.
Suppose $J_{\widehat R}$ is differentiable along the segment from $\theta$
to $\theta+u$, and its gradient is $L$-Lipschitz on that segment.
For any valid certificate $\varepsilon_{\mathrm{credit}}\ge
\|g_L^{(\lambda)}-\nabla J_{\widehat R}(\theta)\|$,
\begin{equation}
\begin{aligned}
a\big(J^*(\theta+u)-J^*(\theta)\big)\ge{}&\big(g_L^{(\lambda)}\big)^\top u
-\varepsilon_{\mathrm{credit}}\|u\|\\
&-\frac L2\|u\|^2-2\varepsilon_{\mathrm{out}}.
\end{aligned}
\label{pr:gain}
\end{equation}
The displacement $u$ is the actual parameter change and may result from
any number of optimizer steps.
\end{lemma}
\begin{proof}
By the fundamental theorem of calculus along the segment,
\begin{align*}
J_{\widehat R}(\theta+u)-J_{\widehat R}(\theta)
&=\nabla J_{\widehat R}(\theta)^\top u
+\int_0^1[\nabla J_{\widehat R}(\theta+tu)
          -\nabla J_{\widehat R}(\theta)]^\top u\,dt\\
&\ge\nabla J_{\widehat R}(\theta)^\top u
-\int_0^1 Lt\|u\|^2\,dt\\
&\ge\big(g_L^{(\lambda)}\big)^\top u-\varepsilon_{\mathrm{credit}}\|u\|
-\frac L2\|u\|^2.
\end{align*}
The last line uses Cauchy--Schwarz. Writing
$\rho(\vartheta)=J_{\widehat R}(\vartheta)-aJ^*(\vartheta)-b$, we have
\begin{align*}
a\big(J^*(\theta+u)-J^*(\theta)\big)
&=J_{\widehat R}(\theta+u)-J_{\widehat R}(\theta)
-\rho(\theta+u)+\rho(\theta)\\
&\ge J_{\widehat R}(\theta+u)-J_{\widehat R}(\theta)
-2\varepsilon_{\mathrm{out}}.
\end{align*}
Combining the inequalities proves Equation~\eqref{pr:gain}.
\end{proof}
A positive right-hand side certifies a positive terminal-return gain because
$a>0$. Endpoint-specific residual bounds can replace
$2\varepsilon_{\mathrm{out}}$ by their sum. If $u$ is random, the statement
holds for each realized update satisfying the stated conditions; no
independence between $u$ and the training data is used.

\subsection{General spaces and episode boundaries}\label{app:regularity}
The same proofs apply to general measurable state and action spaces if
sums are replaced by integrals, functions are measurable and the following
regularity conditions hold. Policy distributions must admit differentiable
densities on a parameter-independent feasible support, satisfy
$\mathbb E[z_{i,h}\mid \mathbf{S}_h]=0$, and allow differentiation under the trajectory
expectation. A sufficient condition is bounded frozen $q_i,v_i,V$ together
with a common trajectory reference measure and an integrable local envelope
for the norm of the trajectory-density derivative. Require also the
integrability of the score-weighted return and advantage terms, and
$\mathbb E\sum_{h,i}\|z_{i,h}\|^2<\infty$ for the Fisher matrix. Spans become
supremum minus infimum, and total variation is the usual total variation of
probability measures. The midpoint argument, Markov recursions and finite
backward sums remain unchanged. Finite spans and finite
$\mathcal C_\pi$ give finite certificates. History states may yield
$\kappa=1$, which is already allowed.

All derivatives hold the learned reward model and advantage-computation
critic fixed. Complete episodes use the terminal conditions stated above.
A zero-boundary rollout cutoff instead defines a truncated-return target;
full-episode and bootstrapped targets retain their corresponding boundary
terms.

\section{Signal floors for structured terminal outcomes}\label{credit:signal-appendix}
Boolean influence analysis supplies the first-level energy and classical tribes family \citep{odonnell2014analysis}. Biased read-once composition also appears in Fourier-entropy/total-influence analysis \citep{odonnell2013composition}. Our signal result lower-bounds first-level energy with a sharp certificate-width coefficient.

\paragraph{Proof roadmap.}
The scalar core, sparse-AND and all-mean composition inequalities lead to
the read-once theorem. The width-two correction handles intermediate
means outside the sparse regime. The one-coalition theorem is a separate
extension; the final corollary connects the floor to calibrated actor
directions.

This appendix proves Theorem~\ref{credit:signal-main} and an extension with
one additional overlapping coalition. Throughout, $A_i$ are independent
Bernoulli actions in a one-decision task, $\rho\le p_i<1$, and
$0<\rho\le1/2$. The outcome $Y$ is a nonconstant monotone Boolean function
of these actions, with mean $\mu=\E Y$. A \emph{minimal winning certificate}
is an inclusion-minimal set of actors whose activation guarantees $Y=1$;
its maximum size is the certificate width $k$. A monotone AND/OR formula
is \emph{read-once} when each physical actor occurs at most once and no
Boolean constants are included. Irrelevant actors may be omitted.

The population individual-credit energy is
\begin{equation}\label{credit:energy-identity}
\mathcal C(Y)=\sum_i\operatorname{Var}(\E[Y\mid A_i])
=\sum_i p_i(1-p_i)(\partial_{p_i}\mu)^2.
\end{equation}
Indeed, conditioning on a Bernoulli coordinate gives
$\partial_{p_i}\mu=\E[Y\mid A_i=1]-\E[Y\mid A_i=0]$.
For $0<u<1$, write
\begin{equation}\label{credit:coefficient-definition}
\varphi(u)=(1-u)^2[-\log(1-u)],\qquad
 a_k(\rho)=\frac{k\rho^{2k-1}(1-\rho)}
 {(1-\rho^k)^2[-\log(1-\rho^k)]}.
\end{equation}
All logarithms are natural. The stated full-support assumptions imply
$0<\mu<1$ and make the equality cases below unambiguous.

For separate actor logits $\theta_i=\log[p_i/(1-p_i)]$, the Fisher matrix
is $F=\operatorname{diag}(p_i(1-p_i))$ and
$\partial_{\theta_i}\mu=p_i(1-p_i)\partial_{p_i}\mu$. Thus
\begin{equation}\label{credit:fisher-energy}
(\nabla_\theta\mu)^\top F^{-1}\nabla_\theta\mu=\mathcal C(Y).
\end{equation}
This identity concerns the true terminal outcome and independently
parameterized logits. The proof below does not assume that fitted local
reward heads equal the conditional expectations in
\eqref{credit:energy-identity}.

\subsection{Exact composition identities}
For a Boolean event $U$ on independent physical actors, write its mean as $u$ and its credit energy as $C_U$. If events $U,V$ use disjoint actor sets, differentiation of their product mean gives
\begin{equation}\label{credit:eq:AND}
C_{U\wedge V}=v^2 C_U+u^2 C_V.
\end{equation}
For independent events $Y_1,\ldots,Y_m$, with means $\mu_j$ and energies $C_j$, their OR has mean $v$ satisfying
\begin{equation}\label{credit:eq:OR}
1-v=\prod_j(1-\mu_j),\qquad
\frac{C_{\vee_jY_j}}{(1-v)^2}
=\sum_j\frac{C_j}{(1-\mu_j)^2}.
\end{equation}
Indeed a derivative belonging to component $j$ is multiplied by the failure probabilities of all other components. Summing its squared contribution proves the identity.
Consequently, if $C_j\ge a\varphi(\mu_j)$ for a common $a$, then
\[
\frac{C_{\vee_jY_j}}{(1-v)^2}
\ge a\sum_j[-\log(1-\mu_j)]
=a[-\log(1-v)].
\]
Equality holds exactly when every component bound is an equality. These identities require disjoint physical actor sets; they do not replace a shared actor by independent copies.

For a singleton $Y=A_i$, put $t=-\log(1-p_i)$. Then
\begin{equation}\label{credit:eq:singleton}
\frac{p_i(1-p_i)}{\varphi(p_i)}
=\frac{e^t-1}{t}=\int_0^1 e^{st}\,ds\ge a_1(\rho),
\end{equation}
with equality exactly when $p_i=\rho$, since the ratio increases strictly
with $p_i$.

For $m$ disjoint coalitions, each containing $k$ actors at probability
$\rho$, set $u=\rho^k$ and $q=1-\mu=(1-u)^m$. A single coalition has
energy $k\rho^{2k-1}(1-\rho)$, so the OR identity gives
\begin{equation}\label{credit:eq:disjoint-equality}
\frac{\mathcal C}{q^2}
=\frac{mk\rho^{2k-1}(1-\rho)}{(1-u)^2}
=a_k(\rho)[-\log q].
\end{equation}
This family establishes sharpness of the coefficient, including the
sparse-success case whenever $\mu\le1/2$.

\subsection{Scalar inequalities}
For $t>0$, set
\[
B(t)=(e^t-1)^2[-\log(1-e^{-t})],\qquad B(0)=0.
\]
Thus $B(-\log u)=\varphi(u)/u^2$.

For the elementary comparisons below, put $y=(x-1)/(x+1)$ and define
\[
\ell_-(x)=2\sum_{m=0}^{3}\frac{y^{2m+1}}{2m+1},
\qquad
\ell_+(x)=\ell_-(x)+\frac{2y^9}{9(1-y^2)}.
\]
The positive logarithm series and its geometric tail give
\begin{equation}\label{credit:eq:ro-logbounds}
\ell_-(x)<\log x<\ell_+(x),\qquad x>1.
\end{equation}
Thus the comparisons below reduce to fixed-degree polynomial inequalities,
without decimal approximations.

\begin{lemma}\label{credit:lem:B}
For $t>0$, $B'(t)>0$, $B''(t)>B'(t)$, and $\log B'(t)$ is strictly concave. Moreover,
\[
\frac{d}{dt}\left(\frac{tB'(t)}{B(t)}\right)>0
\qquad\text{for }t\ge\log2.
\]
\end{lemma}
\begin{proof}
The logarithm series gives
\begin{equation}\label{credit:eq:Bseries}
B(t)=e^t-\frac32+
\sum_{n=1}^{\infty}\frac{2e^{-nt}}{n(n+1)(n+2)}.
\end{equation}
The series and its derivatives converge locally uniformly on $t>0$. Hence
\[
B''(t)-B'(t)=\sum_{n=1}^{\infty}\frac{2e^{-nt}}{n+2}>0.
\]
Also $B'(t)=e^t[1-A(t)]$, where
\[
A(t)=\sum_{m=2}^{\infty}\frac{2e^{-mt}}{m(m+1)}.
\]
The coefficients of $A$ sum to one, so $0<A(t)<1$ and $A''(t)>0$. This proves $B'>0$ and
\[
(\log B')''(t)
=-\frac{A''(t)}{1-A(t)}
-\frac{[A'(t)]^2}{[1-A(t)]^2}<0.
\]

It remains to prove the restricted elasticity claim. Put $p=e^{-t}\le1/2$, $L=-\log(1-p)$, and
\[
\frac{B'(t)}{B(t)}=H(p)=\frac{2-p/L}{1-p},\qquad N(p)=2-p/L.
\]
We first establish $N'(p)/N(p)\le2/3$. Since
$N'=[p/(1-p)-L]/L^2$, this is equivalent to
\[
4L^2+(3-2p)L-\frac{3p}{1-p}\ge0.
\]
The elementary bound $L\ge2p/(2-p)$ follows by differentiating the difference. The preceding expression is increasing in positive $L$ and is therefore at least
\[
\frac{p^2(2-p-4p^2)}{(2-p)^2(1-p)}>0,
\qquad 0<p\le\tfrac12.
\]
Consequently
\[
\frac{H'(p)}{H(p)}=\frac{N'(p)}{N(p)}+\frac1{1-p}\le\frac83.
\]
For $E(t)=tB'(t)/B(t)=tH(e^{-t})$, the inequality $te^{-t}\le1/e$ now gives
\[
E'(t)=H(p)-tpH'(p)
\ge H(p)\left(1-\frac8{3e}\right)>0.
\]
Here $e>8/3$ follows, for example, from its positive power series.
\end{proof}

\begin{lemma}\label{credit:lem:coeff}
For fixed $0<\rho<1$, $a_j(\rho)$ decreases strictly with positive integer $j$.
\end{lemma}
\begin{proof}
Let $T=-\log\rho$ and $K=(1-\rho)/\rho$. Then
\[
a_j(\rho)=\frac{jK}{B(jT)}.
\]
For all $t>0$, put $p=e^{-t}$ and $L=-\log(1-p)$. Since $L>p$,
\[
\frac{B'(t)}{B(t)}=\frac{2-p/L}{1-p}>\frac1{1-p}.
\]
Also $t=-\log p>1-p$, so $tB'(t)/B(t)>1$. Thus $B(t)/t$ increases strictly, proving the assertion. This argument does not assert elasticity monotonicity below $\log2$.
\end{proof}

\begin{lemma}[Core composition]\label{credit:lem:core}
Let $c,l\ge1$ be integers, $0<\rho\le1/2$, $\rho^c\le u\le1$, and $\rho^l\le v<1$. Then
\begin{equation}\label{credit:eq:core}
u^2\left[v^2c(u^{-1/c}-1)+a_l(\rho)\varphi(v)\right]
\ge a_{c+l}(\rho)\varphi(uv).
\end{equation}
Equality holds exactly when $u=\rho^c$ and $v=\rho^l$.
\end{lemma}
\begin{proof}
Write $T=-\log\rho\ge\log2$, $K=e^T-1$, $z=u^{-1/c}\in[1,e^T]$, and $s=-\log v\in(0,lT]$. Dividing by $u^2v^2$, the desired inequality is
\[
F(z,s):=c(z-1)+a_lB(s)-a_{c+l}B(c\log z+s)\ge0.
\]
All coefficients here use the same $\rho$. By Lemma~\ref{credit:lem:B},
\[
\partial_z^2F
=-\frac{a_{c+l}c}{z^2}
\left[cB''(c\log z+s)-B'(c\log z+s)\right]<0.
\]
Thus it suffices to check the two endpoints in $z$. At $z=1$,
$F(1,s)=(a_l-a_{c+l})B(s)>0$ by Lemma~\ref{credit:lem:coeff}.

At $z=e^T$, write
\[
G(s)=cK+a_lB(s)-a_{c+l}B(cT+s),\qquad
R(s)=\frac{B'(cT+s)}{B'(s)}.
\]
Strict concavity of $\log B'$ makes $R$ strictly decreasing. For $s\le lT$,
\[
\frac{a_{c+l}}{a_l}R(s)
\ge\frac{a_{c+l}}{a_l}R(lT)
=\frac{E((c+l)T)}{E(lT)}>1,
\]
where $E(t)=tB'(t)/B(t)$ and both arguments are at least $\log2$. Hence
$G'(s)=B'(s)[a_l-a_{c+l}R(s)]<0$.
The coefficient formula gives $G(lT)=cK+lK-(c+l)K=0$, so $G(s)\ge0$.
Concavity of $F$ proves the inequality. The first endpoint is strictly positive, the second has its unique zero at $s=lT$, and $F$ is strictly concave in $z$. Its only zero is therefore $(z,s)=(e^T,lT)$, giving the asserted equality criterion.
\end{proof}

\subsection{AND composition for a sparse parent}

The next result permits both factors to be general independent component
tasks, provided their individual signal bounds already hold. They need not
be single conjunctions.

\begin{lemma}[Sparse AND composition]\label{credit:thm:ANDsharp}
For integers $c,l\ge1$, $0<\rho\le1/2$, and
\[
\rho^c\le u<1,\qquad \rho^l\le v<1,\qquad uv\le\tfrac12,
\]
one has
\[
\boxed{a_c(\rho)v^2\varphi(u)+a_l(\rho)u^2\varphi(v)
\ge a_{c+l}(\rho)\varphi(uv).}
\]
Equality holds exactly when $u=\rho^c$ and $v=\rho^l$.
\end{lemma}

The proof uses strict subadditivity and a uniform separation between
successive coefficients. Both follow from the same scalar function $B$.

\begin{lemma}\label{credit:lem:hsubadd}
Let $h(t)=B'(t)/B''(t)$. If $t,s>0$ and $t+s\ge\log2$, then
$h(t)+h(s)>h(t+s)$.
\end{lemma}
\begin{proof}
Differentiating the locally uniformly convergent series~\eqref{credit:eq:Bseries},
\[
B''''(t)=e^t+\sum_{n\ge1}\frac{2n^3e^{-nt}}{(n+1)(n+2)}>0.
\]
Hence $B'''$ increases strictly. Lemma~\ref{credit:lem:B} gives
$0<h<1$ and $h'>0$. For $d(t)=h(t)-t$,
\[
d'(t)=-\frac{B'(t)B'''(t)}{[B''(t)]^2}.
\]
Thus $d$ can only increase and then decrease, so its minimum on any
compact interval occurs at an endpoint. Since $B'(t)\to0$ and
$B''(t)\ge e^t$ as $t\downarrow0$, its continuous extension has $d(0)=0$.

Choose the split
\[
L=\log2,\qquad a=\log(6/5),\qquad b=\log(5/3),\qquad a+b=L,
\]
and set $\delta=\min\{d(a),d(b)\}$. We establish the separations
\[
B'''(b)>0,\qquad \delta>0,\qquad d(L)<2\delta.
\]
To see them directly, for $p=e^{-t}$ and $\ell=-\log(1-p)$ write
\[
H(p,z)=\frac{(1-p)(2z-p)}{(4-2p)z-3p},\qquad h(t)=H(p,\ell).
\]
This rational function decreases in $z$ on the intervals used below.
Set
\[
\begin{aligned}
D_-&=\min\{H(5/6,\ell_+(6))-\ell_+(6/5),\\
&\hspace{40mm}H(3/5,\ell_+(5/2))-\ell_+(5/3)\},\\
D_+&=H(1/2,\ell_-(2))-\ell_-(2).
\end{aligned}
\]
The explicit polynomials defining $\ell_\pm$ give
$0<D_+<2D_-$ after clearing their positive denominators.
Equation~\eqref{credit:eq:ro-logbounds} therefore yields
$\delta>D_->D_+/2>d(L)/2$.
Differentiation also gives
\[
B'''(t)=
\frac{(8-2p)\ell-p(7-5p)/(1-p)}{p^2}.
\]
At $p=3/5$, replacing $\ell$ by $\ell_-(5/2)$ leaves a positive numerator,
again a rational polynomial comparison. Hence $B'''(b)>0$.

It follows that $d$ decreases and $h'<1$ on $[b,\infty)$.
Also $d\ge0$ on $[0,b]$ and $d\ge\delta$ on $[a,b]$.
The lower logarithm bound gives $b>1/2$.
If $t,s\ge b$, then $h(t)+h(s)\ge2h(b)>1>h(t+s)$.
If $t\ge b\ge s>0$, integration gives
$h(t+s)-h(t)<s\le h(s)$; the other ordering is symmetric.
Otherwise $t,s<b$ and $t+s\ge L$ put both arguments in $[a,b]$, so
\[
h(t)+h(s)\ge t+s+2\delta
>t+s+d(L)\ge t+s+d(t+s)=h(t+s).
\]
This covers every case.
\end{proof}

\begin{lemma}\label{credit:lem:coefficientgap}
There is a universal constant $\vartheta$ such that, for every integer
$j\ge2$ and $0<\rho\le1/2$,
\[
\frac{a_{j-1}(\rho)}{a_j(\rho)}\ge\vartheta>\frac32,
\qquad
2\vartheta B\!\left(\frac{\log2}{2}\right)>B(\log2).
\]
\end{lemma}
\begin{proof}
Write $L=\log2$ and $T=-\log\rho$. Elasticity monotonicity in
Lemma~\ref{credit:lem:B} shows that each coefficient ratio increases with
$T\ge L$. Define
\[
\vartheta=\min\left\{\frac53,\;
\min_{2\le j\le5}\frac{(j-1)B(jL)}{jB((j-1)L)}\right\}.
\]
For $j\ge6$, $B'/B>1$ implies
\[
\frac{a_{j-1}}{a_j}
=\frac{j-1}{j}\frac{B(jT)}{B((j-1)T)}
>2\frac{j-1}{j}\ge\frac53.
\]
The desired ratio bound follows by definition for the remaining indices.

For the two strict separations, put $x_j=2^j/(2^j-1)$.
The identities
\[
B(jL)=(2^j-1)^2\log x_j,\qquad
B(L/2)=(\sqrt2-1)^2\log(2+\sqrt2)
\]
and~\eqref{credit:eq:ro-logbounds} give
\[
\begin{split}
\vartheta\;&\ge
\min\left\{\frac53,\;
\min_{2\le j\le5}
\frac{(j-1)(2^j-1)^2\ell_-(x_j)}
{j(2^{j-1}-1)^2\ell_+(x_{j-1})}\right\}\\
&>
\max\left\{\frac32,\;
\frac{\ell_+(2)}
{2(\sqrt2-1)^2\ell_-(2+\sqrt2)}\right\}
>
\frac{B(L)}{2B(L/2)}.
\end{split}
\]
The middle comparison is a fixed-degree polynomial inequality: clear
positive denominators, reduce powers of $\sqrt2$ using its square, and
compare the resulting rational terms. This proves both strict separations.
\end{proof}

\begin{proof}[Proof of Lemma~\ref{credit:thm:ANDsharp}]
Put $k=c+l$, $T=-\log\rho$, $t=-\log u$, and $s=-\log v$.
Dividing the desired inequality by $u^2v^2$ gives $F(t,s)\ge0$, where
\[
F(t,s)=a_cB(t)+a_lB(s)-a_kB(t+s).
\]
Consider the compact closure
\[
\mathcal D=\{(t,s):0\le t\le cT,\ 0\le s\le lT,
\ t+s\ge\log2\}.
\]
The function is continuous there, with $B(0)=0$.
On $t=0$, $F=(a_l-a_k)B(s)>0$, and similarly for $s=0$.
On $t=cT$, Lemma~\ref{credit:lem:core} at its anchored core mean gives $F\ge0$,
because $a_cB(cT)=c(e^T-1)$; equality requires $s=lT$.
The face $s=lT$ is symmetric. The intersections of these faces with a
zero coordinate are already covered by the strict zero-face bounds.
On $t+s=\log2$, convexity of $B$ and Lemma~\ref{credit:lem:coefficientgap} give
\[
F\ge a_{k-1}[B(t)+B(s)]-a_kB(\log2)
\ge2a_{k-1}B\!\left(\frac{\log2}{2}\right)-a_kB(\log2)>0.
\]

At an interior stationary point, set $z=t+s$ and
\[
a_cB'(t)=a_lB'(s)=a_kB'(z)=M>0,\qquad r=B''/B'=1/h.
\]
The Hessian is
\[
M\begin{pmatrix}
r(t)-r(z)&-r(z)\\
-r(z)&r(s)-r(z)
\end{pmatrix},
\]
whose determinant is
\[
M^2r(t)r(s)r(z)[h(z)-h(t)-h(s)]<0
\]
by Lemma~\ref{credit:lem:hsubadd}. Thus no interior stationary point can be a
local minimum. The nonnegative boundary values imply $F\ge0$ throughout
$\mathcal D$. An interior zero would also be a local minimum and is
excluded. The boundary analysis leaves only $(t,s)=(cT,lT)$ for equality;
substitution verifies this point.
\end{proof}

\subsection{An all-mean profile for the induction}\label{credit:sec:readonce}

We now remove the restriction on the means of intermediate AND nodes.
The necessary correction affects only width two above success probability
one half. Throughout this section put
\[
L=\log2,\qquad T=-\log\rho\ge L,\qquad K=e^T-1,
\qquad a_j=\frac{jK}{B(jT)},
\]
and define
\begin{align}
b_1&=a_1,& b_2&=\min(a_2,a_1/2),& b_j&=a_j\quad(j\ge3),
\label{credit:eq:ro-b}\\
\kappa_j(u)&=
\begin{cases}
b_2,&j=2\text{ and }u>1/2,\\
a_j,&\text{otherwise}.
\end{cases}\label{credit:eq:ro-profile}
\end{align}
The dependence on the common exploration floor $\rho$ is suppressed.

\paragraph{Why width two needs a correction.}
Let $Y$ be the AND of two independent ORs, each containing $m$ actors
at probability $\rho$. With $\varepsilon=(1-\rho)^m$, direct use of
the composition identities gives
\[
\mu=(1-\varepsilon)^2,\qquad
\mathcal C=\frac{2m\rho}{1-\rho}
\varepsilon^2(1-\varepsilon)^2,\qquad
\frac{\mathcal C}{\varphi(\mu)}\longrightarrow\frac{a_1(\rho)}2.
\]
At $\rho=1/2$, this limit is strictly below $a_2(1/2)$: the comparison
is equivalent to $2^{14}<3^9$. Thus the uncorrected all-mean assertion
fails even for read-once width-two tasks. Conversely, ORs of sufficiently
many disjoint width-two conjunctions have mean above one half and attain
$a_2$ exactly. Any coefficient valid uniformly over nonsparse width-two
tasks must therefore be at most $\min(a_2,a_1/2)$.

\begin{lemma}[All-mean AND composition]\label{credit:thm:profileAND}
For positive integers $c,l$, $0<\rho\le1/2$, and
$\rho^c\le u<1$, $\rho^l\le v<1$,
\[
\boxed{\kappa_c(u)v^2\varphi(u)+\kappa_l(v)u^2\varphi(v)
\ge\kappa_{c+l}(uv)\varphi(uv).}
\]
Equality holds exactly when $u=\rho^c$ and $v=\rho^l$.
\end{lemma}

\subsubsection{Quadratic comparison and corrected coefficients}

\begin{lemma}\label{credit:lem:ro-quadratic}
The function $B(t)/t^2$ decreases strictly on $0<t\le L$. Moreover,
\[
\frac{a_1}{a_3}>3,\qquad
\frac{B'(2L)}{B'(L)}<\frac94.
\]
\end{lemma}
\begin{proof}
Put $q=1-e^{-t}\le1/2$ and $\ell=-\log q$.
The inequality $tB'(t)/B(t)<2$ is equivalent to
\[
2\ell(t-q)<t(1-q).
\]
Choose $\xi=2-L^{-1}$. The elementary bounds $2/3<L<3/4$ give
$1/2<\xi<2/3$. The function
\[
D(q)=\frac{q}{1-\xi q}+\log(1-q)
\]
vanishes at both endpoints of $[0,1/2]$, and its derivative has the sign
of $q[(2\xi-1)-\xi^2q]$. Its minimum is therefore at an endpoint.
Consequently $t\le q/(1-\xi q)$, and
\[
2\ell(t-q)\le2\xi q(-\log q)t
\le\frac{2\xi}{e}t<\frac12t\le(1-q)t.
\]
Here $e>8/3$ follows from its positive power series.

Next, $a_1/a_3=B(3T)/(3B(T))$ increases with $T\ge L$ by
Lemma~\ref{credit:lem:B}. At $T=L$,
\[
B(3L)=49\log(8/7)>9L=9B(L),
\]
where $49\ell_-(8/7)>9\ell_+(2)$ follows from the explicit polynomial
bounds~\eqref{credit:eq:ro-logbounds}. This proves the coefficient claim.
Finally $B'(L)=4L-1$ and $B'(2L)=24\log(4/3)-3$.
The same bounds give
$32\ell_+(4/3)-12\ell_-(2)<1$, hence $4B'(2L)<9B'(L)$.
\end{proof}

\begin{lemma}[Harmonic closure]\label{credit:lem:ro-harmonic}
For all positive integers $c,l$,
\[
\frac1{b_{c+l}}\ge\frac1{b_c}+\frac1{b_l}.
\]
In particular, if $t,s>0$ and $z=t+s\le L$, then
\begin{equation}\label{credit:eq:ro-smallz}
b_cB(t)+b_lB(s)>b_{c+l}B(z).
\end{equation}
\end{lemma}
\begin{proof}
Lemma~\ref{credit:lem:ro-quadratic} gives $a_1>3a_3$. Induction using
Lemma~\ref{credit:lem:coefficientgap} gives $a_1>ja_j$ for every $j\ge3$.
Thus $b_j=\min(a_j,a_1/j)$ for all $j\ge1$.

Put $d_j=1/b_j$. Then $d_2\ge2d_1$, while
\[
\frac{d_3}{d_2}=\frac{b_2}{a_3}
=\min\left(\frac{a_2}{a_3},\frac{a_1}{2a_3}\right)>\frac32.
\]
For $j\ge4$, $d_j/d_{j-1}\ge\vartheta>3/2$.
If $c=l=1$, the claim is the bound on $d_2$.
If one index is one and the other is $m\ge2$, then
$d_{m+1}>(3/2)d_m\ge d_m+d_1$.
If $c\ge l\ge2$, then
$d_{c+l}>(3/2)^l d_c>2d_c\ge d_c+d_l$.
These cases prove the reciprocal inequality.

By Lemma~\ref{credit:lem:ro-quadratic} and Cauchy--Schwarz,
\[
\begin{aligned}
b_cB(t)+b_lB(s)
&>\frac{B(z)}{z^2}(b_ct^2+b_ls^2)\\
&\ge\frac{B(z)}{d_c+d_l}
\ge b_{c+l}B(z).
\end{aligned}
\]
The first inequality is strict because $0<t,s<z\le L$.
\end{proof}

\subsubsection{Shifted and anchored boundary inequalities}

\begin{lemma}\label{credit:lem:ro-shift}
For every $s\ge0$,
\begin{equation}\label{credit:eq:ro-H3}
B(L+s)<\frac32B(L)+3B(s).
\end{equation}
For $0\le s\le L$, this also gives
\begin{equation}\label{credit:eq:ro-symmetricshift}
B(L+s)<\frac94[B(L)+B(s)].
\end{equation}
\end{lemma}
\begin{proof}
Use~\eqref{credit:eq:Bseries}, $B(L)=L$, and $x=e^s\ge1$.
Retaining the first two positive terms gives
\[
\frac32L+3B(s)-B(L+s)
\ge x-3+\frac32L+\frac5{6x}+\frac{11}{48x^2}.
\]
The expression $x+5/(6x)+11/(48x^2)$ is strictly greater than two.
Indeed, multiplying its difference from two by $48x^2$ and putting
$y=x-1\ge0$ gives
\[
48x^3-96x^2+40x+11
=48(y-1/12)^2+8/3+48y^3>0.
\]
The displayed gap is thus greater than $-1+(3/2)L>0$.
The series extends continuously to $s=0$.
If $s\le L$, then $B(s)\le B(L)$ and
$(3/2)B(L)+3B(s)\le(9/4)[B(L)+B(s)]$, proving the second claim.
\end{proof}

\begin{lemma}[Interface faces]\label{credit:lem:ro-interface}
For $l\ge1$, $k=l+2$, $T\ge L$, and $s\ge0$,
\begin{equation}\label{credit:eq:ro-interface}
b_2B(L)+b_lB(s)-a_kB(L+s)>0.
\end{equation}
\end{lemma}
\begin{proof}
For $l=1$, $b_2/a_3>3/2$ and $a_1/a_3>3$, so
\eqref{credit:eq:ro-H3} applies directly.
For $l\ge2$, both $b_2/a_k$ and $b_l/a_k$ exceed
$(3/2)\vartheta>9/4$.
For the first ratio, combine $b_2/a_3>3/2$,
$a_3/a_4\ge\vartheta$, and coefficient monotonicity.
For $l=2$ the second ratio is identical; for $l\ge3$,
$a_l/a_{l+2}\ge\vartheta^2>(3/2)\vartheta$.
Thus \eqref{credit:eq:ro-symmetricshift} gives positivity on $0\le s\le L$.

For $s\ge L$, strict log-concavity of $B'$ and
Lemma~\ref{credit:lem:ro-quadratic} give
\[
\frac{B'(L+s)}{B'(s)}
\le\frac{B'(2L)}{B'(L)}
<\frac94<\frac32\vartheta<\frac{b_l}{a_k}.
\]
Hence the left side of~\eqref{credit:eq:ro-interface} increases strictly
on $[L,\infty)$ from an already positive value.
\end{proof}

\begin{lemma}[Anchored upper faces]\label{credit:lem:ro-anchored}
For $l\ge1$, $k=l+2$, $T\ge L$, and $0\le t\le L$,
\begin{equation}\label{credit:eq:ro-anchored}
b_2B(t)+a_lB(lT)-a_kB(lT+t)>0.
\end{equation}
\end{lemma}
\begin{proof}
Divide the expression by $a_k$, obtaining
\[
H_T(t)=\frac{b_2}{a_k}B(t)+\frac{l}{k}B(kT)-B(lT+t).
\]
The ratio $b_2/a_k$ is the minimum of $a_2/a_k$ and $a_1/(2a_k)$.
Each increases strictly with $T$: its logarithmic derivative is a positive
difference of elasticities divided by $T$, by Lemma~\ref{credit:lem:B}.
The derivative of the last two terms is
$l[B'(kT)-B'(lT+t)]>0$, since $2T>t$ and $B''>0$.
Thus $H_T(t)$ increases with $T$, and its minimum occurs at $T=L$.

Temporarily denote coefficients at $T=L$ by $a_j^*,b_2^*$.
The bound $\log(1+x)<x$ gives $a_k^*>k/(2^k-1)$.
Differentiating~\eqref{credit:eq:Bseries} gives $B'(r)>e^r-1$ for $r>0$:
the coefficients in its subtracted exponential series sum to one.
Therefore
\[
a_k^*B'((k-1)L)
>\frac{k(2^{k-1}-1)}{2^k-1}.
\]
The rational factor increases with $k\ge3$. At $k=3$, exceeding
$2-(2L)^{-1}$ is equivalent to $L<7/10$; this follows from
$\ell_+(2)<7/10$ and~\eqref{credit:eq:ro-logbounds}. Meanwhile,
\[
b_2^*B'(L)\le\frac{4L-1}{2L}=2-\frac1{2L}.
\]
Consequently $a_k^*B'((k-1)L)>b_2^*B'(L)$.
The ratio $B'(lL+t)/B'(t)$ decreases with $t$, so
\[
H_L'(t)=\frac{b_2^*}{a_k^*}B'(t)-B'(lL+t)<0
\qquad(0<t\le L).
\]
Consequently $H_L(t)\ge H_L(L)>0$. The last strict inequality follows
from Lemma~\ref{credit:lem:ro-interface} at $s=lL$, using $a_l\ge b_l$.
Continuity includes $t=0$.
\end{proof}

\subsubsection{Proof of all-mean AND composition}

\begin{proof}[Proof of Lemma~\ref{credit:thm:profileAND}]
Put $t=-\log u$, $s=-\log v$, and $z=t+s$.
If $z<L$, every width-two coefficient involved is in its corrected regime,
and~\eqref{credit:eq:ro-smallz} proves the result strictly.
If $z\ge L$ and neither input uses the correction, the assertion is
Lemma~\ref{credit:thm:ANDsharp}. This includes $c=l=1$, for which the parent
coefficient returns to $a_2$ at $z=L$.

It remains to consider a corrected input. By symmetry let $c=2$ and
$0<t<L$. First suppose $l\ne2$. On the compact region
\[
0\le t\le L,\qquad0\le s\le lT,\qquad t+s\ge L,
\]
consider the continuous expression
\[
F(t,s)=b_2B(t)+a_lB(s)-a_{l+2}B(t+s).
\]
All its boundary faces are strictly positive. On a zero-coordinate face,
use coefficient monotonicity and $b_2>a_3$. The face $t=L$ is covered by
Lemma~\ref{credit:lem:ro-interface}, with $a_l\ge b_l$; the face $s=lT$ is
Lemma~\ref{credit:lem:ro-anchored}. On the positive part of $t+s=L$, use
\eqref{credit:eq:ro-smallz}, $b_l\le a_l$ and $b_{l+2}=a_{l+2}$.
The endpoints of this diagonal are already covered by the zero faces.

At an interior stationary point the three first-derivative terms equal
some $M>0$. With $r=B''/B'=1/h$, the Hessian determinant is
\[
M^2r(t)r(s)r(z)[h(z)-h(t)-h(s)]<0
\]
by Lemma~\ref{credit:lem:hsubadd}, exactly as in the proof of
Lemma~\ref{credit:thm:ANDsharp}. There is no interior local minimum.
Compactness and the strictly positive boundary therefore give $F>0$.

If $l=2$, split the region at $s=L$. On its lower piece use $b_2$ for
both inputs; on its upper piece use $b_2$ for the first and $a_2$ for
the second. The stationary-point argument is unchanged. All exterior faces
were just treated, and the new interface $s=L$ is positive by
Lemma~\ref{credit:lem:ro-interface}, with the factors interchanged. Using $a_2$
instead of $b_2$ can only increase that boundary value. Thus both pieces
are strictly positive. At an input mean exactly one half,
\eqref{credit:eq:ro-profile} uses $a_2$; the boundary arguments deliberately
allowed the smaller $b_2$, so this jump leaves no missing case.

The small-$z$ and corrected-input regions are strict. In the remaining
region, Lemma~\ref{credit:thm:ANDsharp} gives equality precisely at
$(t,s)=(cT,lT)$. These inputs use the uncorrected coefficients and do
attain equality, completing both directions.
\end{proof}

\subsubsection{The read-once theorem}

A monotone AND/OR formula is \emph{read-once} if each physical actor appears
in at most one leaf of the expression. This permits intersecting successful
coalitions after expansion, and arbitrary alternation of
AND and OR. It does not permit copying an actor to make two component
tasks independent.

\begin{theorem}[Sharp signal bound for read-once tasks]\label{credit:signal-profile}
Let $Y$ be a finite nonconstant read-once monotone AND/OR formula, with no
Boolean constants, on independent Bernoulli actors satisfying
$\rho\le p_i<1$ and $0<\rho\le1/2$. Let $k$ be the maximum size of a
minimal winning certificate. At every $0<\mu=\E Y<1$,
\[
\boxed{\mathcal C(Y)\ge\kappa_k(\mu)\varphi(\mu).}
\]
In particular, if $\mu\le1/2$, then
$\mathcal C(Y)\ge a_k(\rho)\varphi(\mu)$ with the coefficient in Theorem~\ref{credit:signal-main}. In that sparse regime, equality holds if and only if the
minimal winning coalitions are disjoint, each has size $k$, and every
relevant actor has probability $\rho$.
\end{theorem}
\begin{proof}
The singleton bound is~\eqref{credit:eq:singleton}, valid at every
probability $p\ge\rho$.
At an AND node, child actors are disjoint and their certificate widths
add. A child of width at most $c$ has a winning certificate with probability
at least $\rho^c$, so its mean is at least that value.
The exact identity~\eqref{credit:eq:AND}, followed by
Lemma~\ref{credit:thm:profileAND}, propagates the profile bound through the
node. Higher-arity ANDs may be grouped into binary nodes because that
lemma applies at all intermediate means.

For an OR node, the width $k$ is the maximum child width, and
\eqref{credit:eq:OR} gives
\[
\frac{C_{\vee_jY_j}}{(1-\mu)^2}
=\sum_j\frac{C_{Y_j}}{(1-\mu_j)^2},\qquad
-\log(1-\mu)=\sum_j[-\log(1-\mu_j)].
\]
Every child coefficient is at least the parent coefficient.
For $k\ge3$, this uses $b_2>a_3\ge a_k$ and ordinary coefficient
monotonicity. For $k=2$ and $\mu\le1/2$, every child mean is at most
$\mu$, so no child uses the correction. For $k=2$ and $\mu>1/2$,
the parent coefficient $b_2$ is the minimum possible child coefficient.
The case $k=1$ is immediate. The two identities therefore propagate the
bound through OR. Induction proves the all-mean assertion.

Now assume $\mu\le1/2$ and equality. At an equality AND node,
Lemma~\ref{credit:thm:profileAND} forces the child means to be $\rho^c$
and $\rho^l$. For any monotone event of width at most $c$, the presence
of one certificate already gives mean at least $\rho^c$.
Equality in this lower bound forces a single certificate of exactly $c$
actors, all with probability $\rho$: a shorter certificate or a larger
probability increases its probability, and an additional minimal
certificate adds a positive-probability success assignment under the
full-support product law. Thus an equality AND node is a single
conjunction at the exploration floor.

At an equality OR node, its child means are also at most one half.
Strict coefficient monotonicity forces every child width to equal the
parent width, and~\eqref{credit:eq:OR} forces every child bound to be an equality.
Recursive application gives exactly disjoint size-$k$ conjunctions with
all relevant probabilities equal to $\rho$. Conversely, equation~\eqref{credit:eq:disjoint-equality} gives equality
for that family.
\end{proof}

\begin{proof}[Proof of Theorem~\ref{credit:signal-main}]
Let $j\le k$ be the actual maximum certificate width. For $\mu\le1/2$,
the profile theorem gives $\mathcal C\ge a_j(\rho)\varphi(\mu)$.
Lemma~\ref{credit:lem:coeff} gives $a_j(\rho)\ge a_k(\rho)$, strictly
when $j<k$. The preceding theorem therefore gives the claimed bound;
equality requires $j=k$ and its stated disjoint-coalition equality case.
\end{proof}

\subsection{One overlapping coalition, with optional fresh actors}\label{credit:sec:onecrossfresh}

A single insertion into the disjoint equality family permits an explicit
form of actor reuse. The added coalition may also contain actors outside
the original task.

\begin{theorem}[One additional coalition]\label{credit:signal-onecross}
Start with $m\ge1$ disjoint coalitions of $k\ge1$ independent actors
each, all acting with probability $\rho$, where $0<\rho\le1/2$.
Add one distinct coalition of $k$ actors. It selects $r_j$ actors from
each of $h$ old blocks, with
\[
1\le r_j<k,\qquad R=\sum_{j=1}^h r_j\le k,
\]
and uses $k-R$ fresh independent actors, also with probability $\rho$.
The case $h=R=0$ is allowed. Let $Y$ indicate completion of at least one old or new coalition.
At every $0<\mu=\E Y<1$,
\[
\boxed{\mathcal C\ge a_k(\rho)(1-\mu)^2[-\log(1-\mu)].}
\]
Equality holds exactly when $R=0$. Every permitted overlap, $R>0$,
gives strict inequality.
\end{theorem}

The condition $r_j<k$ excludes reproducing an old coalition. At $R=k$
all added actors come from the old system; at $R=0$ the new coalition
is independent of it.

\begin{proof}
Put
\[
u=\rho^k,\qquad h_0=-\log(1-u),\qquad
e_0=\frac{\rho^{2k-1}(1-\rho)}{(1-u)^2},\qquad
a_k=\frac{ke_0}{h_0}.
\]
For failure probability $q=1-\mu$, write
\[
\mathcal E=\frac{\mathcal C}{q^2}
=\sum_i\rho(1-\rho)(-\partial_{p_i}\log q)^2,
\qquad \mathcal L=-\log q.
\]
Derivatives are taken in the independent physical probabilities and
then evaluated at $p_i=\rho$.
By~\eqref{credit:eq:disjoint-equality}, the old disjoint system satisfies
$\mathcal E_{\rm old}=a_k\mathcal L_{\rm old}$.
Conditional on its failure, the new coalition is complete with probability
\[
\delta=ux,\qquad
x=\prod_{j=1}^h\frac{1-\rho^{k-r_j}}{1-u}\in(0,1].
\]
The selected old actors contribute $\rho^R$, while the fresh actors
contribute $\rho^{k-R}$. Define
\[
\alpha=R/k,\qquad D=1-ux,\qquad W=\frac{ux}{D}.
\]
Then $\Delta\mathcal L=-\log(1-ux)$.

Every old actor has old log-failure derivative
$g=\rho^{k-1}/(1-u)$. In a hit block, let $a$ and $b$ be the products
of its selected and unselected probabilities. At general physical
probabilities, this block contributes $a(1-b)/(1-ab)$ to $\delta$.
Thus
\[
\partial_{p_i}\log\delta=
\begin{cases}
\displaystyle\frac{1}{p_i(1-ab)},& i\text{ selected},\\[4pt]
\displaystyle-\frac{b(1-a)}{p_i(1-b)(1-ab)},& i\text{ unselected}.
\end{cases}
\]
Since $q_{\rm new}=q_{\rm old}(1-\delta)$, differentiating before
setting $p_i=\rho$ gives the new log-failure derivatives
\[
-\partial_{p_i}\log q_{\rm new}=
\begin{cases}
g(1+W/u),& i\text{ selected from an old block},\\
g(1-Wf_j),& i\text{ unselected in hit block }j,\\
g,& i\text{ in an untouched old block},\\
W/\rho,& i\text{ fresh},
\end{cases}
\qquad
f_j=\frac{\rho^{-r_j}-1}{1-\rho^{k-r_j}}.
\]
A fresh actor has old derivative zero. Define
\[
\mathcal A=\sum_j(k-r_j)f_j,\qquad
\mathcal B=\sum_j(k-r_j)f_j^2\ge0.
\]
There are $R$ selected old actors and $k-R$ fresh actors, and
$\rho(1-\rho)g^2=e_0$. Summing the squared derivative changes over
all physical actors yields the exact identity
\begin{equation}\label{credit:eq:fresh-energy}
\frac{\Delta\mathcal E}{e_0}
=2W\left(\frac{R}{u}-\mathcal A\right)
+W^2\left[\frac{R}{u^2}+\mathcal B+
\frac{(k-R)(1-u)^2}{u^2}\right].
\end{equation}
For integers $r,s\ge1$ with $r+s=k$, geometric sums give
\[
1-\rho^r\le r(1-\rho),\qquad
s\rho^{s-1}(1-\rho)\le1-\rho^s,
\]
and hence
\[
u\,s\frac{\rho^{-r}-1}{1-\rho^s}
=\frac{s\rho^s(1-\rho^r)}{1-\rho^s}\le r\rho.
\]
Summing over hit blocks proves $u\mathcal A\le\rho R$.

A second geometric comparison gives $x\ge1-\alpha$.
For $r+s=k$ with $r,s\ge1$, put
\[
g_r=1-\frac{1-\rho^s}{1-u}
=\frac{\sum_{a=s}^{k-1}\rho^a}{\sum_{a=0}^{k-1}\rho^a}
\le\frac{r}{k}.
\]
The numerator consists of the $r$ smallest terms of the decreasing
sequence $1,\rho,\ldots,\rho^{k-1}$. Since $0\le g_r<1$,
\begin{equation}\label{credit:eq:fresh-x}
x=\prod_j(1-g_{r_j})
\ge1-\sum_jg_{r_j}\ge1-R/k=1-\alpha.
\end{equation}

Use $\rho\le1/2$, $\mathcal B\ge0$, and $W=ux/D$ in
\eqref{credit:eq:fresh-energy} to obtain
\begin{equation}\label{credit:eq:fresh-lower}
\frac{\Delta\mathcal E}{e_0}
\ge k\left[
\frac{\alpha x}{D}
+\frac{x^2\{\alpha+(1-\alpha)(1-u)^2\}}{D^2}\right].
\end{equation}
After factoring $kx/D^2$, the numerator inside this expression is
\[
x(1-u)^2+\alpha[1+u(1-u)x].
\]
Its coefficient of $\alpha$ is positive. Substituting
$\alpha\ge1-x$ from~\eqref{credit:eq:fresh-x} and simplifying yields
\begin{equation}\label{credit:eq:fresh-mainbound}
\frac{\Delta\mathcal E}{e_0}
\ge kx\left[1+\frac{ux(1-x)}{(1-ux)^2}\right]\ge kx.
\end{equation}
Convexity of $z\mapsto-\log(1-z)$, whose value at zero is zero, gives
\[
\Delta\mathcal L=-\log(1-ux)
\le x[-\log(1-u)]=xh_0.
\]
Combining this with~\eqref{credit:eq:fresh-mainbound} proves
$\Delta\mathcal E\ge a_k\Delta\mathcal L$.
Adding the old equality and multiplying by $q_{\rm new}^2$ establishes
the asserted signal bound.

If $R>0$, at least one $g_{r_j}$ is strictly positive and every factor
in $x$ is positive. Hence $0<x<1$, and the first lower bound in
\eqref{credit:eq:fresh-mainbound} is strictly greater than $kx$.
If $R=0$, the new coalition is independent of every old coalition and
the disjoint-block computation gives equality. This proves both directions.
\end{proof}

The argument also gives an explicit increase of the normalized deficit:
\[
\Delta(\mathcal E-a_k\mathcal L)
\ge ke_0\left[
\frac{ux^2(1-x)}{(1-ux)^2}
+x-\frac{-\log(1-ux)}{h_0}\right].
\]
Both terms in brackets are nonnegative, and the first is positive for
every permitted overlap. This establishes stability of the disjoint
equality family under one such insertion, including instances without
a read-once representation. It does not justify starting from an already
overlapping system, performing repeated insertions, or changing to
heterogeneous actor probabilities.

\paragraph{An instance without a read-once representation.}
Take $k=m=2$, old coalitions $\{1,2\},\{3,4\}$ and new coalition
$\{2,3\}$. The resulting success event is
\[
A_1A_2\lor A_2A_3\lor A_3A_4.
\]
Its minimal true certificate $\{2,3\}$ and minimal false certificate
$\{A_2=0,A_3=0\}$ intersect in two actors. In a monotone read-once
formula, every minimal true certificate meets every minimal false
certificate in exactly one actor. This follows by induction: at an OR
node the true certificate chooses one child while the false certificate
includes one from each child; at an AND node the roles reverse.
Thus this instance has no read-once representation in the physical
actors. At $\rho=1/2$ it has $\mu=1/2$ and $\mathcal C=5/32$, so it
also lies in the sparse-success domain of Theorem~\ref{credit:signal-main}.

The independent-composition arguments require disjoint physical actor
sets; identifying independent copies of the same actor changes the task.
The read-once theorem and the single-insertion theorem therefore do not
establish a bound for arbitrary overlapping coalition families.

\subsection{From signal strength to a calibrated local direction}\label{app:signal-transfer}
Specialize the setting of Theorem~\ref{sf:theorem} to the product-action
model in Section~\ref{sec:signal-floor}. Each actor makes its Bernoulli
decision once, with a separate logit coordinate; subsequent steps have
forced actions and parameter-independent dynamics. The terminal outcome
is the stated function of the initial decisions. The complete trajectory
score then contains only the initial action scores, so
$\mathcal I_\pi=F=\operatorname{diag}(p_i(1-p_i))$ is positive definite.
In particular, the error bound and the signal identity
\eqref{credit:fisher-energy} use the same metric.

\begin{corollary}[Calibrated natural-direction certificate]\label{credit:alignment}
Let the model satisfy Theorems~\ref{sf:theorem}
and~\ref{credit:signal-main} in this common coordinate system. Write
$B_{\rm cred}=(\varepsilon/2)\sum_{t=1}^{H-1}\gamma^t$.
For a frozen scale $a>0$, assume
\[
\|\nabla J_{\widehat R}-a\nabla\mu\|_{F^{-1}}\le\delta_{\rm cal}.
\]
If
\begin{equation}
B_{\rm cred}+\delta_{\rm cal}
<a(1-\mu)\sqrt{a_k(\rho)\log\frac1{1-\mu}},
\label{credit:alignment-main}
\end{equation}
then
$\nabla\mu^\top F^{-1}g_L^{(1)}>0$.
\end{corollary}
\begin{proof}
The triangle inequality and Theorem~\ref{sf:theorem} give
$\|g_L^{(1)}-a\nabla\mu\|_{F^{-1}}\le B_{\rm cred}+\delta_{\rm cal}$.
Hence
\[
\nabla\mu^\top F^{-1}g_L^{(1)}
\ge\|\nabla\mu\|_{F^{-1}}
\left[a\|\nabla\mu\|_{F^{-1}}-B_{\rm cred}-\delta_{\rm cal}\right]>0.
\]
The last inequality uses \eqref{credit:fisher-energy},
Theorem~\ref{credit:signal-main}, and \eqref{credit:alignment-main}.
\end{proof}
The corollary concerns the first-order natural direction $F^{-1}g_L^{(1)}$
at a frozen checkpoint. Its gradient-calibration assumption is distinct
from accurate pair ordering or endpoint return calibration.
It does not certify a finite clipped-PPO displacement. Arbitrary actor
parameter sharing can project away part of this independently parameterized
signal, so the floor is not asserted after such a projection. The
shared-parameter credit-error theorem remains valid in its own stated
setting.

\section{Implementation and benchmark protocol}\label{app:implementation}
\IMAPSmacLiteTable
All benchmarks use terminal-outcome feedback. SMACv2 actors and local reward heads share parameters across homogeneous agents, with agent identity included in the observations. They use recurrent history encoders. MAMuJoCo actors and reward heads are separate across agents; actors and critics use two hidden layers of width 256. Reward-model encoders follow the domain's actor/critic encoder design but have separate parameters and optimizers. Discrete actions are one-hot encoded before the Q head, and continuous actions are concatenated with the observation embedding.

\paragraph{Outcome-comparison buffer.}
The buffer persists across PPO iterations. Binary pairs come from the win and loss groups; continuous-return pairs come from groups separated by the buffer's running median. Only completed episodes enter the comparisons. Equal-return pairs are discarded, and fitting is skipped when no unequal-return pair is available.

\begin{table}[ht]
\centering\small
\caption{Benchmark training settings. The relay uses the separate settings below.}
\label{tab:benchmark_config}
\begin{tabular}{lcc}
\toprule
Setting & SMACv2 & MAMuJoCo\\
\midrule
Actor / critic learning rate & $0.0005$ & $0.0003$\\
Discount / GAE & $0.99$ / $0.95$ & $0.99$ / $0.95$\\
PPO clipping / epochs & $0.2$ / 10 & $0.2$ / 10\\
PPO minibatch size & 256 & 512\\
Entropy coefficient & $0.01$ & $0.01$\\
Parallel environments & 8 & 8\\
Rollout length & 400 & 200\\
Training steps & 10M & 10M\\
Evaluation episodes & 32 & 32\\
Reward-model learning rate & $0.0003$ & $0.0003$\\
Reward-model pair batch & 256 & 512\\
IMAP label-bearing Q updates / iteration & 5 & 5\\
IMAP additional V updates / iteration & 5 & 5\\
Comparison FIFO capacity & 8192 pairs & 8192 pairs\\
Extreme-V temperature & $1.0$ & $1.0$\\
Reward-square coefficient & $0.001$ & $0.001$\\
Reward-model gradient clipping & $0.5$ & $0.5$\\
\bottomrule
\end{tabular}
\end{table}

\paragraph{Terminal feedback and rollout boundaries.}
SMACv2 disables intermediate damage, kill and health rewards. MAMuJoCo retains its task dynamics and cumulative evaluation return, while the learner receives only the complete-episode scalar. Complete episodes generate comparison labels even when an actor rollout ends earlier. Actor GAE uses the actual termination mask and its rollout bootstrap convention; a reset observation is not a successor inside the completed episode. The complete-episode theory diagnostics use zero terminal local values and critic values.

\paragraph{Reward and advantage snapshots.}
The normalized scalar mixer is shared by Q and V. A shared affine transform of the form $(r_i-\mu_R)/\sigma_R$, applied identically to the global reward, preserves $\sum_i\alpha_i r_i=\widehat R$. Each diagnostic stores the raw Q/V outputs, the scalar weights, the normalization parameters and the critic snapshot. Both GAE streams use these same fixed quantities and masks. The residual identity is verified before comparing gradients. If an implementation applies an additional reduction or advantage transform, its resulting vector is evaluated with that transform; the theoretical vector is computed with the stated discounted time weights.

\paragraph{Reward scaling.}
Certificates use the same frozen heads and reward scale as the compared
vectors. Under a shared positive rescaling of $q_i,v_i,V$ by
$1/\sigma_R$, reward-part gradients, compatibility spans and their
certificates scale by $1/\sigma_R$. State-constant reward shifts at fixed
stages do not alter these spans. If a shift or mask varies by state,
compatibility must instead be recomputed from the transformed heads.
The numerical studies here use raw residuals.

For the benchmark controls, the shared affine reward transform is applied
before advantage construction. The objectives use the actor prefactors
in Equation~\eqref{eq:actor} and a unit-prefactor global control.

\paragraph{What each ablation changes.}
SL-MAPPO and IMAP-GA share the global actor readout, terminal comparisons and label-bearing update schedule. SL-MAPPO predicts rewards directly; IMAP-GA uses Q/V residuals and additionally optimizes V, so total optimizer work differs. IMAP-LA retains the Q/V reward model with local actor advantages. Online-IPL extracts policies from learned Q/V values by advantage-weighted behavior cloning. Its extraction family is related to O-MAPL, whose published evaluation uses a separate offline protocol \citep{bui2025omapl}. SparseMAPPO and SparseHAPPO optimize terminal rewards directly.

The unit-prefactor global and $\alpha_i$-weighted local readouts preserve the same one-stage population gradient (Appendix~\ref{app:objective-consistency}). The local coefficient retains its component's weight in the mixed objective; multiplying the global readout by $\alpha_i$ again would square this coefficient. This explains the readout convention, including its common entropy term. Sequential GAE, finite samples, advantage transforms and repeated clipped updates can still produce different optimization behavior. The experiments compare complete configurations, without attributing each gain to an isolated mechanism.

\paragraph{Additional held-out fitting measurements.}
The held-out reward-model diagnostic freezes IMAP-GA and SL-MAPPO after 10M training steps. Per task and seed, both models score the same 2000 pairs drawn from a common held-out pool of 256 trajectories, with trajectories from both policies included. Table~\ref{tab:heldout_reward} reports descriptive means over three seeds. Pairs sharing trajectories are not separate experimental replicates. This diagnostic concerns score ordering and likelihood; the policy-gradient and outcome-calibration measurements are defined separately.
\IMAPHeldoutTable

\begin{figure}[t]
\centering
\includegraphics[width=0.9\linewidth]{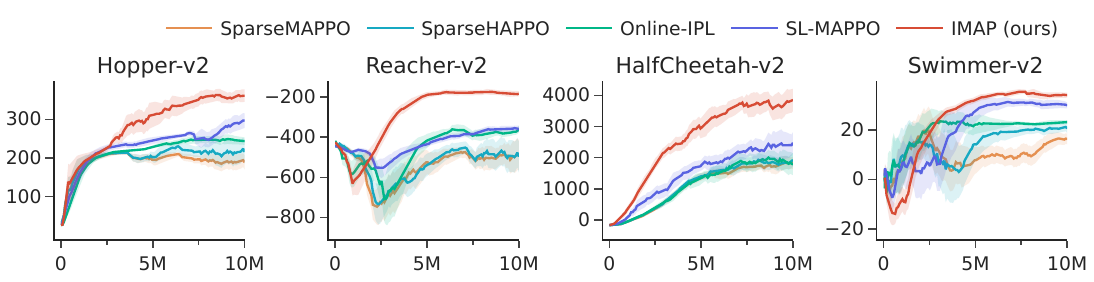}
\caption{MAMuJoCo comparison with the benchmark baselines. IMAP (ours) denotes the terminal-outcome IMAP-Rule/LA configuration; this figure supplements the main-text GA/LA ablation.}
\label{fig:mujoco_all}
\end{figure}

\section{Relay protocol and numerical credit checks}\label{app:toy-protocol}
\label{app:toy-results}
The relay is a finite diagnostic environment for the IMAP construction in Section~\ref{sec:method}. Four independent uniform cue bits determine four correct binary actions. Agents observe stage, current cue, surviving-success status and active role. Alternating ownership assigns successive decisions to different agents; concentrated ownership assigns all four to the first agent. Both conditions retain the same two-agent roster and parameter allocation. Failure changes the progress flag but leaves the decision schedule intact. Terminal reward is one for four correct choices and zero otherwise.

\paragraph{Networks and fitting.}
Actors allocate 32 Bernoulli logits indexed by agent, decision stage, progress and cue. The shared critic allocates 276 scalar values over a common state registry, including unused suffix states in shorter episodes. Observations have 12 coordinates: scaled time, five stage indicators, cue, progress, two role indicators and two identity indicators. Each local Q network has widths $15\!\to\!32\!\to\!32\!\to\!1$ and each local V network $12\!\to\!32\!\to\!32\!\to\!1$, with SiLU activations. Including two mixer logits, the reward model has 6214 parameters. Hidden layers use Xavier-uniform initialization; final output heads and mixer logits start at zero. Forced no-ops have singleton action supports, and terminal local and critic values are zero.

The model uses IMAP's terminal-outcome comparison loss and mixed Extreme-V update, with temperature one and reward-square coefficient $0.001$. Per 64-episode PPO batch, five fitting cycles each use 128 comparison pairs. The comparison FIFO capacity is 8192. Both Q/mixer and V have their own optimizer state. Each 4096-episode run has 320 Q/mixer and 320 V updates. PPO uses ten epochs, learning rate $0.0005$, clipping $0.2$, entropy coefficient $0.01$, value coefficient $0.5$, $\gamma=1$ and $\lambda=0.95$. Reward-model learning rate is also $0.0005$. All gradient norms are clipped at $0.5$.

\paragraph{Complete checkpoint population.}
The numerical collection contains eleven Q/V configurations: GA and LA at delays 0, 16 and 64 with alternating ownership; GA and LA with concentrated ownership at delay 16; and GA with fixed fitting datasets containing 1, 4 or 16 observed successes. Each configuration has seeds 0--9. Four saved checkpoints per run, at batches 16, 32, 48 and 64, give $11\times10\times4=440$ cases. All cases are included regardless of policy success, compatibility error or bound ratio. The fixed-data cases use 512 fitting episodes from a common behavior-policy pool per seed, with nested success/failure samples, then freeze the reward model during policy collection. They enter here as fitted models for numerical verification, not as learning-performance comparisons.

\paragraph{Frozen targets and reductions.}
At each checkpoint, the reward model is frozen after fitting, while the actor and critic are frozen before that batch's PPO update. Local/global rewards use the same model, weights, masks and critic snapshot. No per-agent reward rescaling or advantage centering is used. The actor loss averages the four active decision positions per episode; critic loss averages within episodes and then across episodes. At $\gamma=1$, the implemented actor mean is one quarter of the theorem's unnormalized sum. The reported Fisher-error and continuation checks use the latter. Stored episodes and repeated pair samples are not treated as independent outcome labels.

\paragraph{Exact path calculations.}
Every frozen policy is evaluated on all 256 cue/action paths. Independent
return differentiation and a fixed-target surrogate give the team and
local actor vectors. The trajectory Fisher uses the complete undiscounted
score and its numerically resolved range. All checks use fixed
scale-aware tolerances; near-zero bounds are handled separately from
ratios. No relay checkpoint is excluded. Panel (c) uses each checkpoint's
original error and bound, targeting the learned objective $J_{\widehat R}$.

\paragraph{Controlled decomposition maps.}
The heatmaps use all ten final reference IMAP-LA checkpoints: alternating ownership, delay 16, seeds 0--9, after 4096 collected episodes. Each snapshot fixes the post-fitting reward model and the actor/critic immediately before its final PPO update. No model or policy is refitted during the intervention. The original 440-checkpoint collection remains a separate coverage check; the heatmaps use the fixed reference condition so that delay, ownership and model-fitting regime do not vary within a map.

For each seed, let
\[
s_{\rm QV}=\max\!\left\{
\max_{t,i}\operatorname{span}_{\mathbf{s},a_i}q_{i,t}(o_i(\mathbf{s}),a_i),\;
\max_{t,i}\operatorname{span}_{\mathbf{s}}v_{i,t}(o_i(\mathbf{s}))
\right\},
\]
where each span uses the reachable states and feasible actions at its stage. All ten scales are positive. Generate independent standard-normal entries $\psi_i(\mathbf{s})$ on the relay's public state registry using the fixed per-model seeds in the accompanying numerical record. Center them as $\widetilde\phi_i(\mathbf{s})=\psi_i(\mathbf{s})-\sum_j\alpha_j\psi_j(\mathbf{s})$, set terminal entries to zero, and define
\[
\phi_i(\mathbf{s})=s_{\rm QV}\,
\frac{\widetilde\phi_i(\mathbf{s})}
{\max_{t,j}\operatorname{span}_{\mathbf{s}}\widetilde\phi_{j,t}(\mathbf{s})}.
\]
Every observed denominator is positive. The same potentials and fixed scale are used in both intervention channels and at every grid cell within a seed. The public state is present in each agent's observation, so adding these action-independent functions preserves the local-head information restriction.

Use $\lambda\in\{0,0.05,\ldots,1\}$ and $\zeta\in\{-2,-1.75,\ldots,2\}$. For each of the two channels, ten seeds give $10\times21\times17$ exact evaluations, or 7140 in total. Each heatmap cell is the arithmetic mean of ten responses normalized before averaging. Both maps use one unsaturated linear color scale; cells are displayed without interpolation. No seed, sign, endpoint or high-response cell is excluded. The complete per-seed grids retain their ranges as well as their means. The response is a magnitude, so its color does not encode an improvement direction.

\begin{table}[t]
\centering\small
\caption{Variation across the ten fixed reference models at training $\lambda=0.95$ and unit amplitude $\zeta=1$. Values use the normalized response in Equation~\eqref{eq:toy_response}; SD and range are across seeds, not grid cells.}
\label{tab:heatmap_seed_variation}
\begin{tabular}{lccc}
\toprule
Output shift & Mean & Sample SD & Seed range\\
\midrule
Joint Q/V & 0.00605 & 0.00275 & $[0.00236,\,0.01026]$\\
Q-only & 0.11496 & 0.05218 & $[0.04488,\,0.19488]$\\
\bottomrule
\end{tabular}
\end{table}

\begin{figure}[t]
\centering
\includegraphics[width=\linewidth]{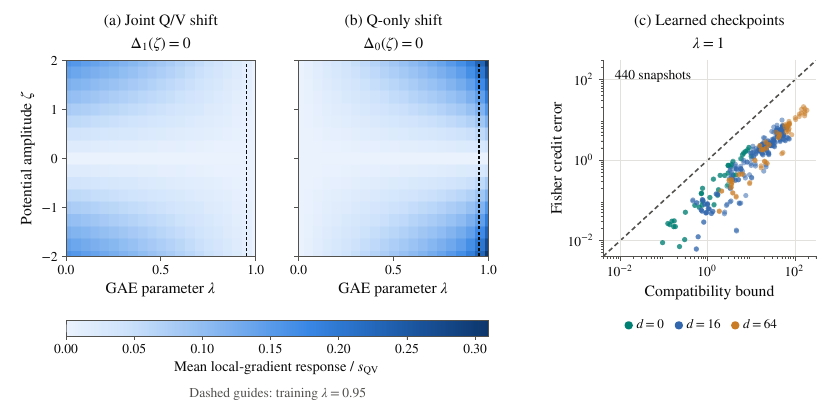}
\caption{Local-credit sensitivity at unchanged mixed reward. (a,b) Mean response $\Delta_\lambda(\zeta)$ over ten frozen models; shared color scale and dashed training $\lambda=0.95$. Joint Q/V shifts preserve compatibility; Q-only shifts change it. (c) Full-return Fisher error versus the Theorem~\ref{sf:theorem} bound for all 440 checkpoints; maximum error/bound ratio $0.2620$, with diagonal denoting equality. All panels use the frozen learned-return objective.}
\label{fig:toy_verification}
\end{figure}

Both channels leave $Q_{\rm tot}$ and $V_{\rm tot}$ unchanged. Consequently the mixed reward, its trajectory score and the reward-fitting losses on fixed data are unchanged, although individual actor directions can move. These are interventions on frozen outputs, not claims that a particular network refit realizes the altered functions. For the joint Q/V shift,
\[
\Delta e_i=0,\qquad \Delta D_i=-\zeta\phi_i;
\]
for the Q-only shift,
\[
\Delta e_i=\zeta\phi_i,\qquad \Delta D_i=0.
\]
Substituting in Equation~\eqref{pr:F} shows that the forcing changes are respectively $-(1-\lambda)\zeta\phi_i$ and $\lambda\zeta\phi_i$, while $\bar e$ remains fixed. This explains the opposite zero-response endpoints in Figure~\ref{fig:toy_verification}. At a fixed $\lambda$, the two response vectors are collinear for this shared-potential construction. The factor of 19 between their magnitudes at $\lambda=0.95$ follows from the coefficients; it is a consistency check, not an empirical advantage of one update.

Independent surrogate differentiation agrees with the score and continuation formulas at the declared numerical tolerances. Both interventions preserve the mixed Q/V functions and reward, their zero-response endpoints are recovered, and no tested full-return or stagewise bound is violated.

The complete simultaneous-channel grid also agrees with the sum of the separate responses. Since these shared-potential channels are collinear, this is an algebraic control rather than an independent mechanism.

For a GAE parameter $\lambda$, the normalized actor response used in
Figure~\ref{fig:toy_verification} and Table~\ref{tab:heatmap_seed_variation} is
\begin{equation}
\Delta_\lambda(\zeta)=
\frac{\|g_L^{(\lambda)}(\zeta)-g_L^{(\lambda)}(0)\|_{\mathcal I_\pi^\dagger}}
{s_{\rm QV}}.
\label{eq:toy_response}
\end{equation}

\paragraph{Delay control.}
A fixed-policy terminal-reward control confirms full-return gradient
invariance across the three delays. In fitted models, however, the
reward-square penalty averages over all transitions, including the suffix.
Different suffixes therefore change this averaging as well as GAE
propagation; they do not isolate exploration difficulty.

Model fitting uses float32 arithmetic; all exact path diagnostics use float64. Stored model, policy and critic snapshots specify the frozen functions used in each check.

\section{Numerical verification of the theory}\label{app:numerics}

\begin{figure}[t]
\centering
\includegraphics[width=0.6\linewidth]{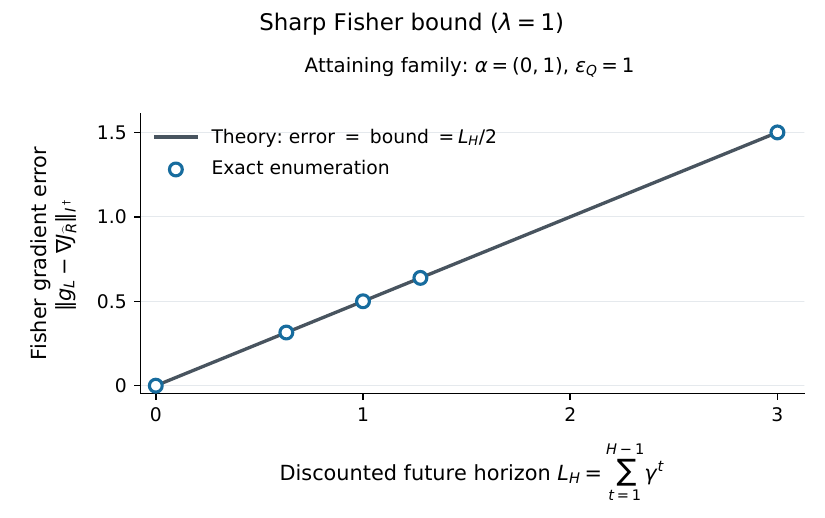}
\caption{The attaining two-agent family at $\lambda=1$, $\alpha=(0,1)$ and $\varepsilon_Q=\varepsilon=1$. Analytic error and bound coincide at $L_H/2$. Six saved finite-model checks occupy five distinct coordinates; strictly positive mixers approach the same constant.}
\label{fig:sharpness}
\end{figure}

These checks use frozen finite models and are separate from benchmark
performance. They cover coupled transitions, shared categorical policy
parameters, observation aliases and explicit attaining constructions.
Independent enumeration and recursive differentiation agree at the
declared numerical tolerances; no tested identity, Fisher-range condition
or sharp bound is violated. Near-zero bounds are checked directly,
without interpreting unstable error/bound ratios.

Figure~\ref{fig:sharpness} verifies numerical attainment in the analytic
family and approach to the same constant with strictly positive mixers.
The proof establishes sharpness; these calculations check its realization.

\paragraph{Reproducibility record.}
The numerical record in the supplementary material specifies the fixed seeds,
spectral cutoffs, scale-aware tolerances, complete audit populations,
fitting-loss ranges and per-case output locations used in Appendices
\ref{app:toy-protocol}--\ref{app:signal-diagnostic}.
All original models and numerical records are retained.

\section{Instantiating the signal-to-error condition}\label{app:signal-diagnostic}
This finite-population diagnostic evaluates both sides of
\eqref{credit:alignment-main} on fitted Q/V heads in the setting of
Theorem~\ref{credit:signal-main}. Actors are fixed throughout; no policy
learning or PPO improvement is measured.

\paragraph{Tasks and fixed data.}
Four actors independently choose binary actions once. The next state
publicly records their joint action, all subsequent actions are forced
no-ops, and the binary outcome is delivered at termination. With $H=2$
and $\gamma=1$, the four tasks are
\[
A_0A_1A_2A_3,\quad
A_0A_1\lor A_2A_3,\quad
(A_0\lor A_1)(A_2\lor A_3),\quad
A_0\big(A_1\lor A_2A_3\big),
\]
with maximum certificate widths $4,2,2,3$, respectively.
Each training dataset contains all 16 joint-action trajectories under
the complete uniform-behavior population. Only their terminal outcomes
supply labels. Every unequal-outcome pair is used, oriented winner first.

We fit all four tasks with seeds $0$--$4$, giving 20 reward models.
Each actor has a Q network $15\!\to\!32\!\to\!32\!\to\!1$ and a V network
$12\!\to\!32\!\to\!32\!\to\!1$, with SiLU hidden layers, plus four shared
mixer logits; the total is 12428 parameters. The 12 observation coordinates
contain a three-way stage indicator, four public action bits (zero before
the decision), two identity bits, a constant, and two unused zeros.
Q additionally receives a three-way action encoding, including the no-op.
Hidden weights use Xavier-uniform initialization; biases, output layers
and mixer logits start at zero.

Each fit performs 512 Q-then-V cycles with the objectives
\eqref{eq:rewardloss}--\eqref{eq:extremev}, temperature one,
reward-square coefficient $0.001$, learning rate $0.0005$ and gradient
clipping $0.5$. Q/mixer and V use separate Adam states with
$(0.9,0.999)$ moments and $10^{-8}$ denominator constant.
V is frozen in the Q update; Q and mixer are frozen in the V update.
The reward-square penalty averages the complete positive/negative pair
population with its trajectory multiplicities. The V loss uses all 16
trajectories and both nonterminal transitions uniformly. This specifies
the fitting distribution independently of the diagnostic policies.

\paragraph{Complete checkpoint and policy grid.}
Checkpoints at cycles $0,32,128,512$ give 80 snapshots, including
initialization. Each is evaluated at three fixed product policies:
$p_i=0.2$ for all actors, $p_i=0.4$ for all actors, and
$p=(0.25,0.35,0.45,0.55)$. Set $\rho=\min_i p_i$.
All twelve task/policy combinations satisfy $0<\mu\le1/2$, with
$\mu\in[0.0016,0.4096]$. Every one of the 240 cases is retained.

Full trajectory enumeration computes the true gradient $g_Y=\nabla\mu$,
learned-score gradient $g_F=\nabla J_{\widehat R}$, ideal energy
$\mathcal C=g_Y^\top F^{-1}g_Y$, and its lower bound
$\mathcal C_{\rm floor}$ from \eqref{credit:floor-main}.
The full-return local vector is
$(g_L)_i=\alpha_i\E[(A_i-p_i)G_i]$, where $G_i=\sum_h r_{i,h}$.
We use a zero shared critic; full-return gradients are unchanged by the
stage-zero baseline. Since the initial state is fixed and stage one has
forced actions,
\[
\varepsilon=\max_i\operatorname{span}_{\mathbf A}
  [q_{i,1}(\mathbf A,\mathrm{noop})-v_{i,1}(\mathbf A)],
\qquad B_{\rm cred}=\varepsilon/2.
\]

\paragraph{Calibration is measured as a diagnostic.}
The primary scale is the explicitly oracle gradient projection
\[
a_g=\frac{g_F^\top F^{-1}g_Y}{g_Y^\top F^{-1}g_Y},\qquad
\delta_g=\|g_F-a_g g_Y\|_{F^{-1}}.
\]
A second diagnostic uses the return least-squares scale
$a_{\rm out}=\operatorname{Cov}(\widehat F,Y)/\operatorname{Var}(Y)$,
with intercept $\E\widehat F-a_{\rm out}\mu$ and its corresponding
gradient residual. A nonpositive scale is inadmissible.
Both use the true population outcome, and $a_g$ additionally uses its
gradient. Neither is learned from pair ordering or proposed as an
operational calibration estimator.

A case is certified only when
$a\sqrt{\mathcal C_{\rm floor}}-B_{\rm cred}-\delta_{\rm cal}$
exceeds $\varepsilon_{\rm num}[1+|a\sqrt{\mathcal C_{\rm floor}}|
+|B_{\rm cred}+\delta_{\rm cal}|]$.
Here $\varepsilon_{\rm num}$ is the fixed arithmetic tolerance specified
in the accompanying numerical record. We also evaluate the signed alignment
$g_Y^\top F^{-1}g_L$ directly.

\begin{table}[t]
\centering\small
\caption{Complete product-action checkpoint grid. Counts refer to fixed
task/seed/policy cases, not independent statistical replicates.
Certification uses \eqref{credit:alignment-main} at the stated tolerance.}
\label{tab:bridge_checkpoints}
\fitwidth{
\begin{tabular}{rcccc}
\toprule
Fitting cycles & Cases & Positive alignment & Oracle-scale certified & Return-scale certified\\
\midrule
0 & 60 & 0 & 0 & 0\\
32 & 60 & 60 & 0 & 0\\
128 & 60 & 60 & 0 & 0\\
512 & 60 & 60 & 6 & 0\\
\bottomrule
\end{tabular}}
\end{table}

All 60 initialization cases have zero learned directions and inadmissible
zero scales; undefined normalized quantities remain recorded as such.
All 180 nonzero fitted cases have positive natural alignment, but only six
are certified under $a_g$: four AND4 cases and two OR-of-AND cases, all
at cycle 512. No case is certified under $a_{\rm out}$.
Thus the criterion is sufficient and conservative. These results show
limited nonvacuity on learned heads under an oracle calibration premise;
they do not establish a calibration procedure or benchmark-scale benefit.

\paragraph{Which term limits certification?}
Table~\ref{tab:bridge_budget_components} decomposes the normalized budget
for all 60 final base cases. Write $s_g=a_g\sqrt{\mathcal C_{\rm floor}}$;
each case passes only if its own two ratios sum to less than one, subject
to the stated numerical tolerance. The reported medians and ranges
summarize each component separately and must not be added as if they
came from the same case.

\begin{table}[t]
\centering\small
\caption{Budget components at cycle 512, over all five seeds and three
policies per task. Entries are medians [minimum, maximum], not confidence
intervals. Certificates use the pointwise sum; counts repeat the final
row of Table~\ref{tab:bridge_checkpoints}.}
\label{tab:bridge_budget_components}
\fitwidth{
\begin{tabular}{lccc}
\toprule
Task & $B_{\rm cred}/s_g$ & $\delta_g/s_g$ & Certified / 15\\
\midrule
AND4 & $1.155\ [0.543,2.466]$ & $0.059\ [0.034,0.639]$ & 4\\
OR-of-AND & $1.344\ [0.774,1.758]$ & $0.018\ [0.008,0.470]$ & 2\\
AND-of-OR & $2.586\ [1.746,4.128]$ & $0.028\ [0.017,0.515]$ & 0\\
Nested & $1.921\ [1.032,2.447]$ & $0.379\ [0.120,0.500]$ & 0\\
\bottomrule
\end{tabular}}
\end{table}

The credit-error term alone exceeds the unit budget in every AND-of-OR
and nested case. Thus, even zero calibration residual would not certify
those cases with the same credit bound and signal floor. The two simpler
tasks include cases below the unit credit budget, where the calibration
residual must still be accounted for. This describes the existing
sufficient test, not a necessity result or an additional experiment.

\paragraph{A fixed-score intervention grid.}
At every final snapshot and policy, use
$\zeta\in\{-2,-1,-0.5,0,0.5,1,2,4\}$.
At the public stage-one state define
\[
\phi_i(\mathbf A)=-(A_i-p_i)+\sum_j\alpha_j(A_j-p_j).
\]
Its mixture is pointwise zero. Normalize its maximum component span to
the unshifted heads' scale $s_{\rm QV}$, set it to zero at the other
stages, and shift $q_{i,1}$ by $\zeta\phi_i$ with V fixed.
The potentials are frozen before differentiation.
The policy, critic, mixer and neural parameters are unchanged; this is
an intervention on outputs, not a refit or policy update.
The eight amplitudes were fixed before execution, and all 480 cases are
retained.

\begin{table}[t]
\centering\small
\caption{Every prescribed intervention at the final fitting checkpoint.
Each row contains 60 cases. The six certificates at $\zeta=0$ repeat the
six final cases in Table~\ref{tab:bridge_checkpoints}. Return-scale
certification is zero at every amplitude.}
\label{tab:bridge_interventions}
\begin{tabular}{rccc}
\toprule
$\zeta$ & Positive alignment & Negative alignment & Oracle-scale certified\\
\midrule
$-2$ & 60 & 0 & 0\\
$-1$ & 60 & 0 & 0\\
$-0.5$ & 60 & 0 & 0\\
$0$ & 60 & 0 & 6\\
$0.5$ & 60 & 0 & 0\\
$1$ & 57 & 3 & 0\\
$2$ & 1 & 59 & 0\\
$4$ & 0 & 60 & 0\\
\bottomrule
\end{tabular}
\end{table}

\begin{figure}[t]
\centering
\includegraphics[width=\linewidth]{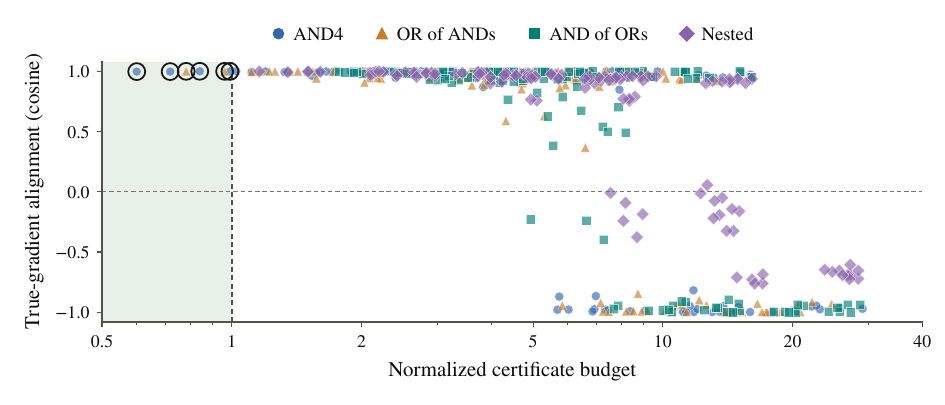}
\caption{All 480 frozen product-action interventions. The horizontal
coordinate is $(B_{\rm cred}+\delta_g)/(a_g\sqrt{\mathcal C_{\rm floor}})$;
the vertical coordinate is the cosine between $g_Y$ and $g_L$ in
$F^{-1}$ geometry. Outlined points are certified and all have positive
alignment. Many points outside the certificate region are also positive:
failure of the sufficient test does not imply failure of the direction.
The calibration scale is oracle, and the six outlined points are the
unshifted controls already counted in Table~\ref{tab:bridge_checkpoints}.}
\label{fig:signal_bridge}
\end{figure}

\begin{samepage}
The interventions preserve the mixed score, reward and fitting objectives
at the declared numerical tolerances. Nevertheless, 122 local directions
become negatively aligned with the true gradient; none is certified.
This distinguishes score fit from local-credit accuracy without implying
that ordinary fitting selects these interventions.
\end{samepage}

\paragraph{Fitting and independent checks.}
Final pair ordering is exact on three tasks; on the AND-of-OR task,
accuracy is at least $0.968$ across the retained seeds. These are
fixed-budget fits, not convergence claims. Final per-task BT and Extreme-V
loss ranges are in the numerical record; its linked complete data also
retain every initialization case.

A separate implementation recomputes the neural outputs, gradients,
spans, losses and certificates. The results agree at the declared
tolerances, with no signal-floor or credit-bound violation and no
certified nonpositive alignment. All snapshots and failed certificates
remain in the complete record.

\end{document}